\documentclass[10pt,twocolumn,letterpaper]{article}

\usepackage{cvpr}
\usepackage{times}
\usepackage{epsfig}
\usepackage{graphicx}
\usepackage{amsmath}
\usepackage{amssymb}
\usepackage{multirow}
\usepackage{booktabs}

\usepackage[pagebackref=true,breaklinks=true,letterpaper=true,colorlinks,citecolor=blue,linkcolor=blue,bookmarks=false]{hyperref}

 \cvprfinalcopy 

\def\cvprPaperID{816} 

\ifcvprfinal\fi
\begin{document}

\title{Learning Dual Convolutional Neural Networks for Low-Level Vision}

\vspace{-4mm}
\author{Jinshan Pan$^{1}$\quad Sifei Liu$^{2}$\quad Deqing Sun$^{2}$ \quad Jiawei Zhang$^{3}$ \quad Yang Liu$^{4}$ \quad Jimmy Ren$^{5}$ \\
Zechao Li$^{1}$ \quad Jinhui Tang$^{1}$  \quad Huchuan Lu$^{4}$ \quad Yu-Wing Tai$^{6}$\quad Ming-Hsuan Yang$^{7}$ \\
$^{1}$Nanjing University of Science and Technology \quad $^{2}$NVIDIA \quad  $^{3}$City University of Hong Kong \\
$^{4}$Dalian University of Technology   \quad $^{5}$SenseTime Research \quad  $^{6}$Tencent Youtu Lab\quad $^{7}$UC Merced\\
{\url{https://sites.google.com/site/jspanhomepage/dualcnn}}\\
}

\maketitle

\begin{figure}[t]\footnotesize
\vspace{-2.3in}
\begin{minipage}{\textwidth}
\begin{center}
\begin{tabular}{cccccc}
\hspace{-2mm}
\includegraphics[width = 0.162\linewidth, height = 0.14\linewidth]{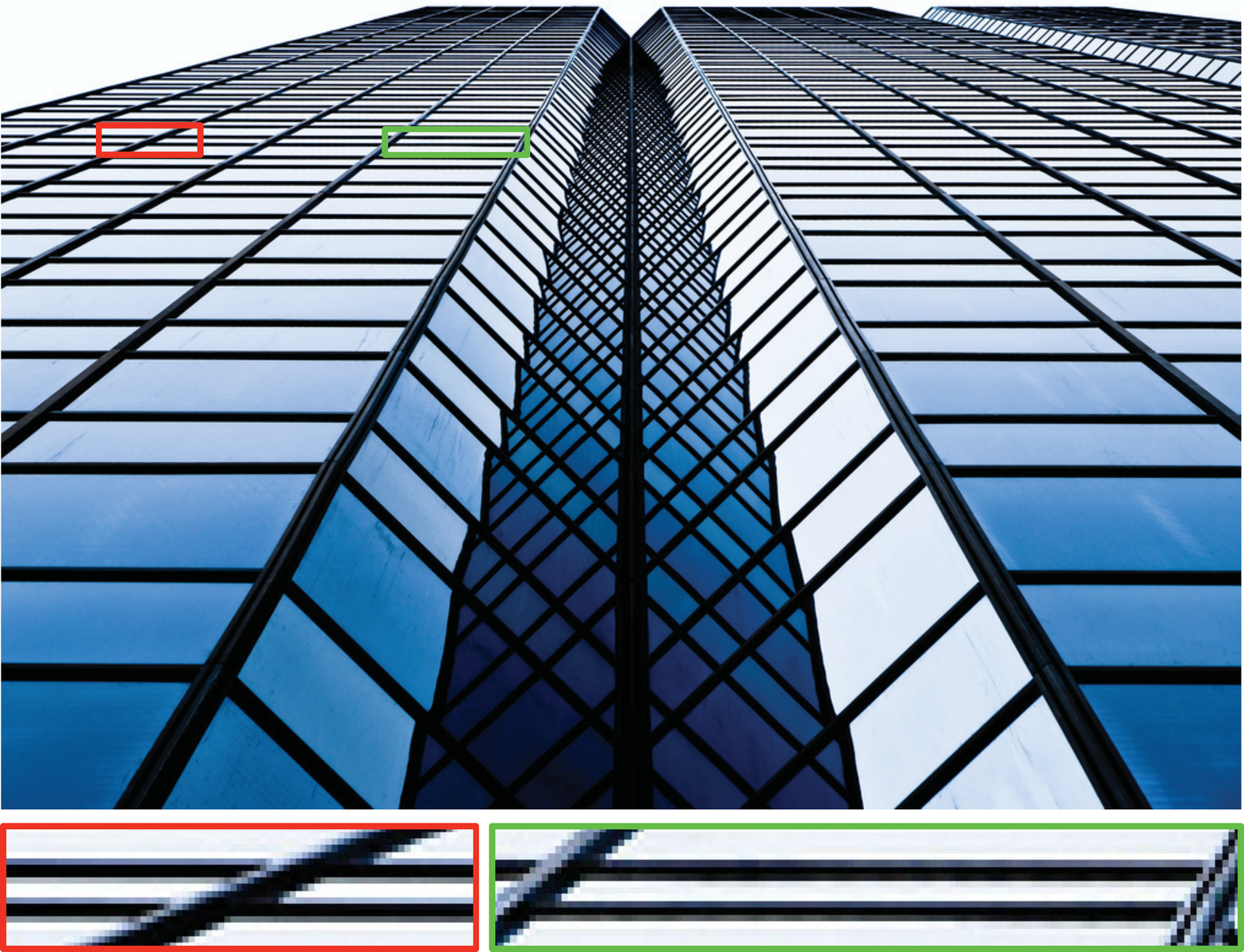}& \hspace{-4mm}
\includegraphics[width = 0.162\linewidth, height = 0.14\linewidth]{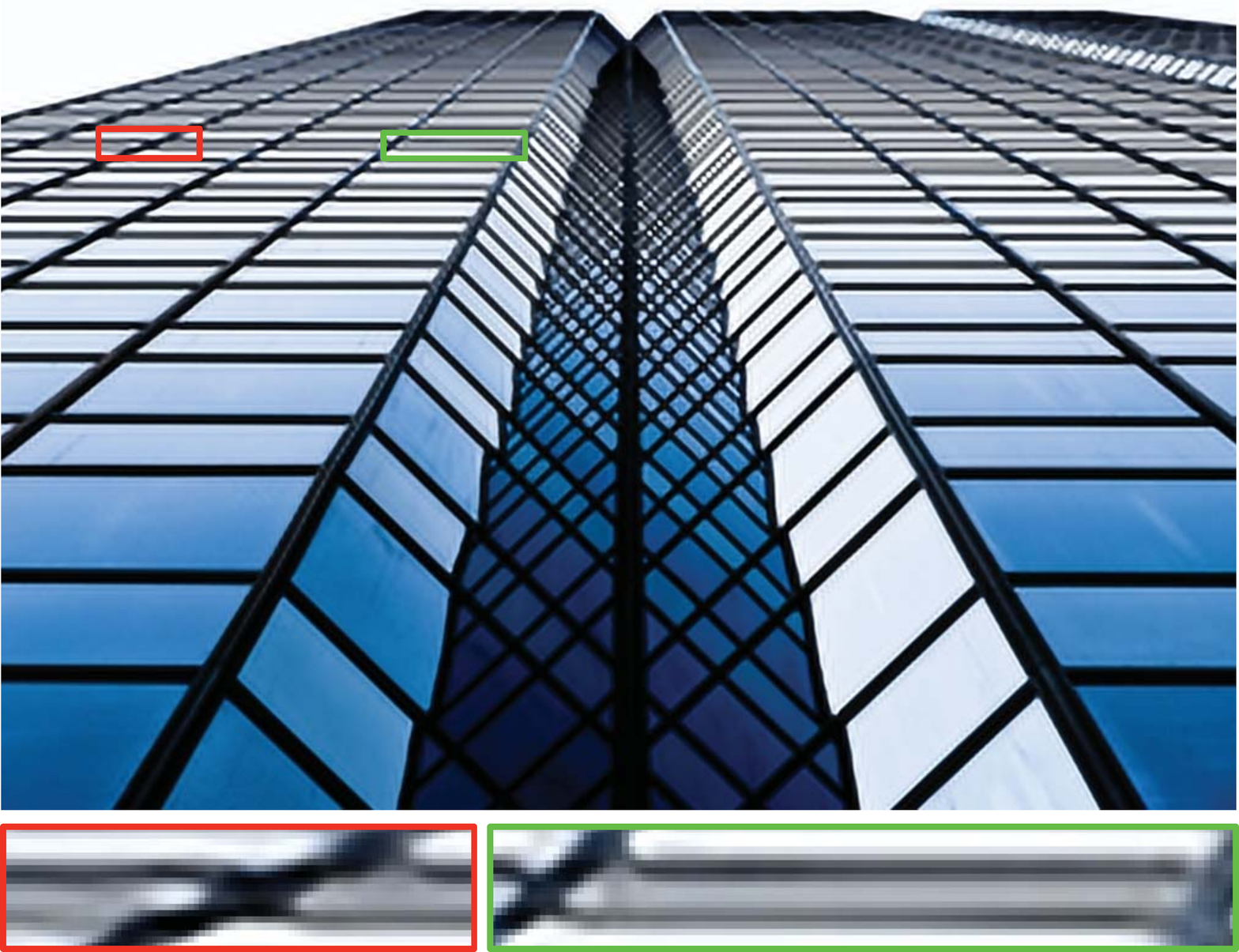}& \hspace{-4mm}
\includegraphics[width = 0.162\linewidth, height = 0.14\linewidth]{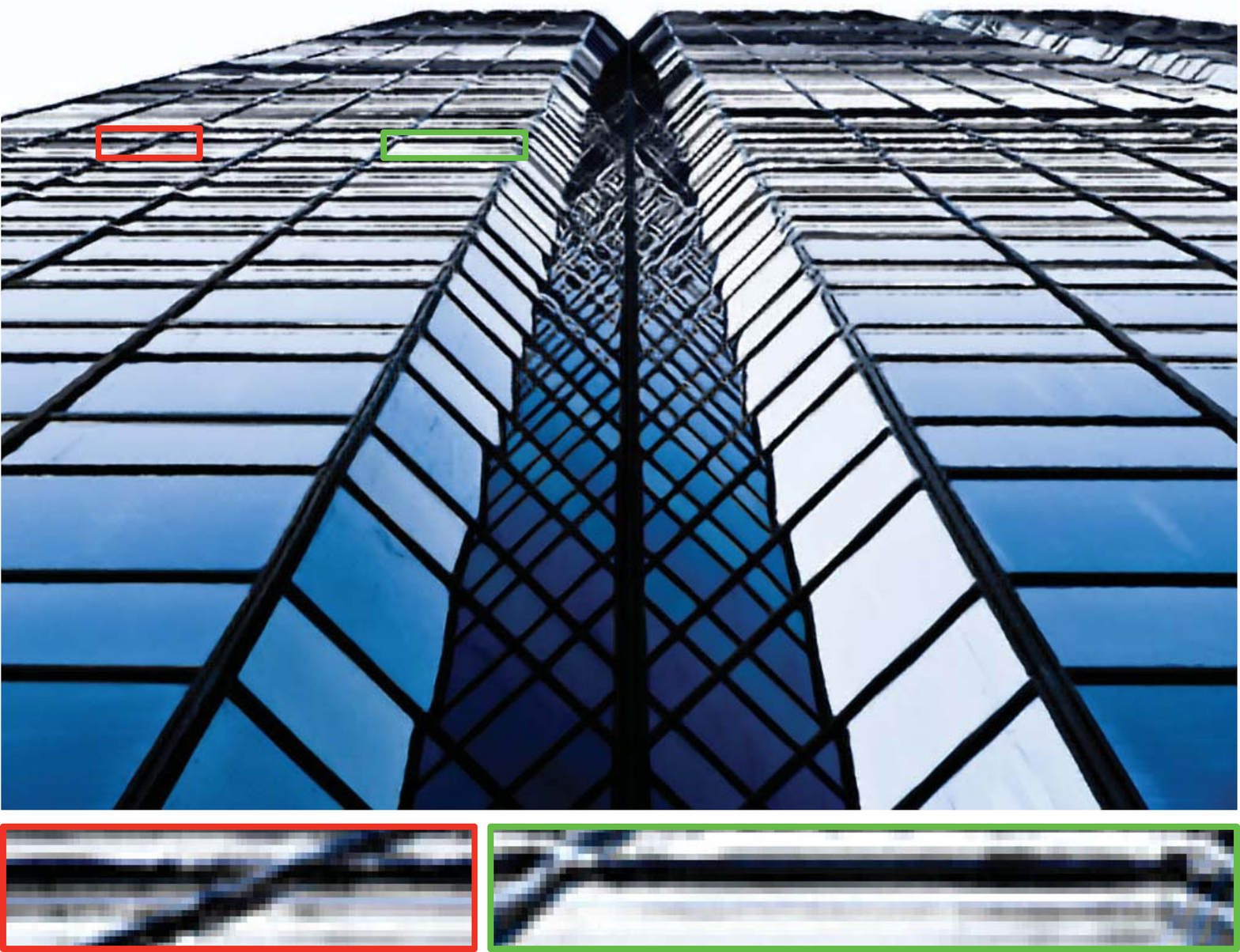} &\hspace{-4mm}
\includegraphics[width = 0.162\linewidth, height = 0.14\linewidth]{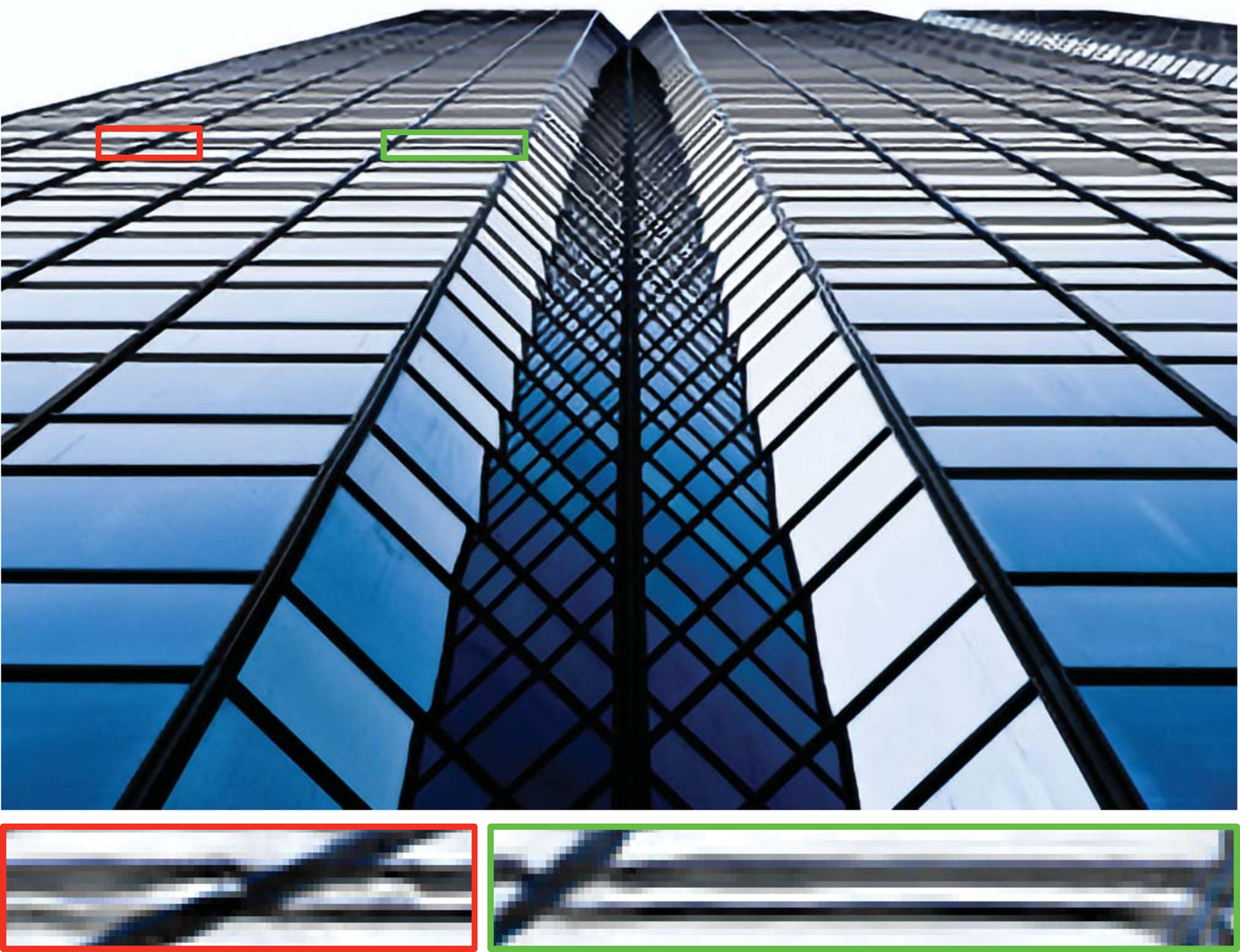}& \hspace{-4mm}
\includegraphics[width = 0.162\linewidth, height = 0.14\linewidth]{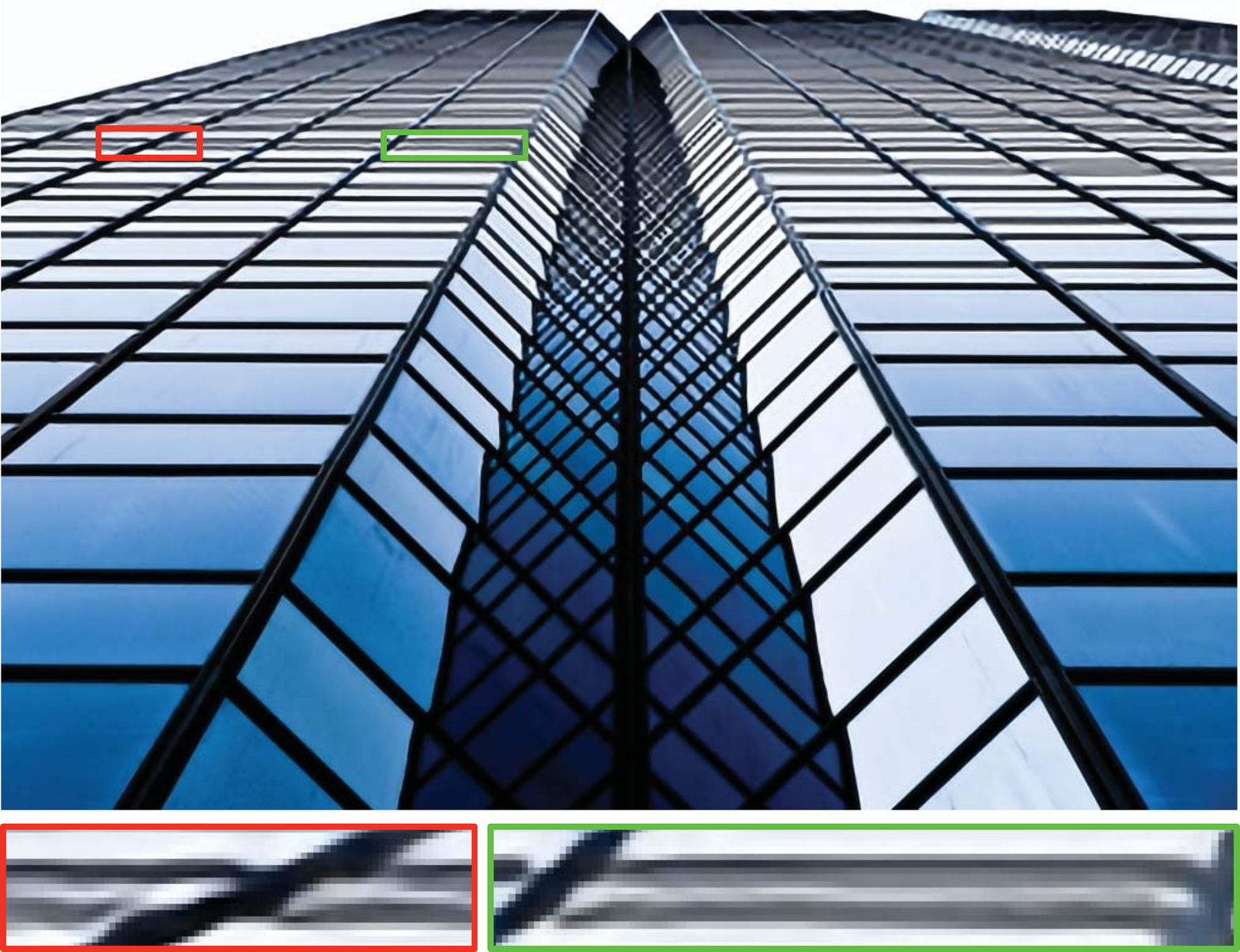}& \hspace{-4mm}
\includegraphics[width = 0.162\linewidth, height = 0.14\linewidth]{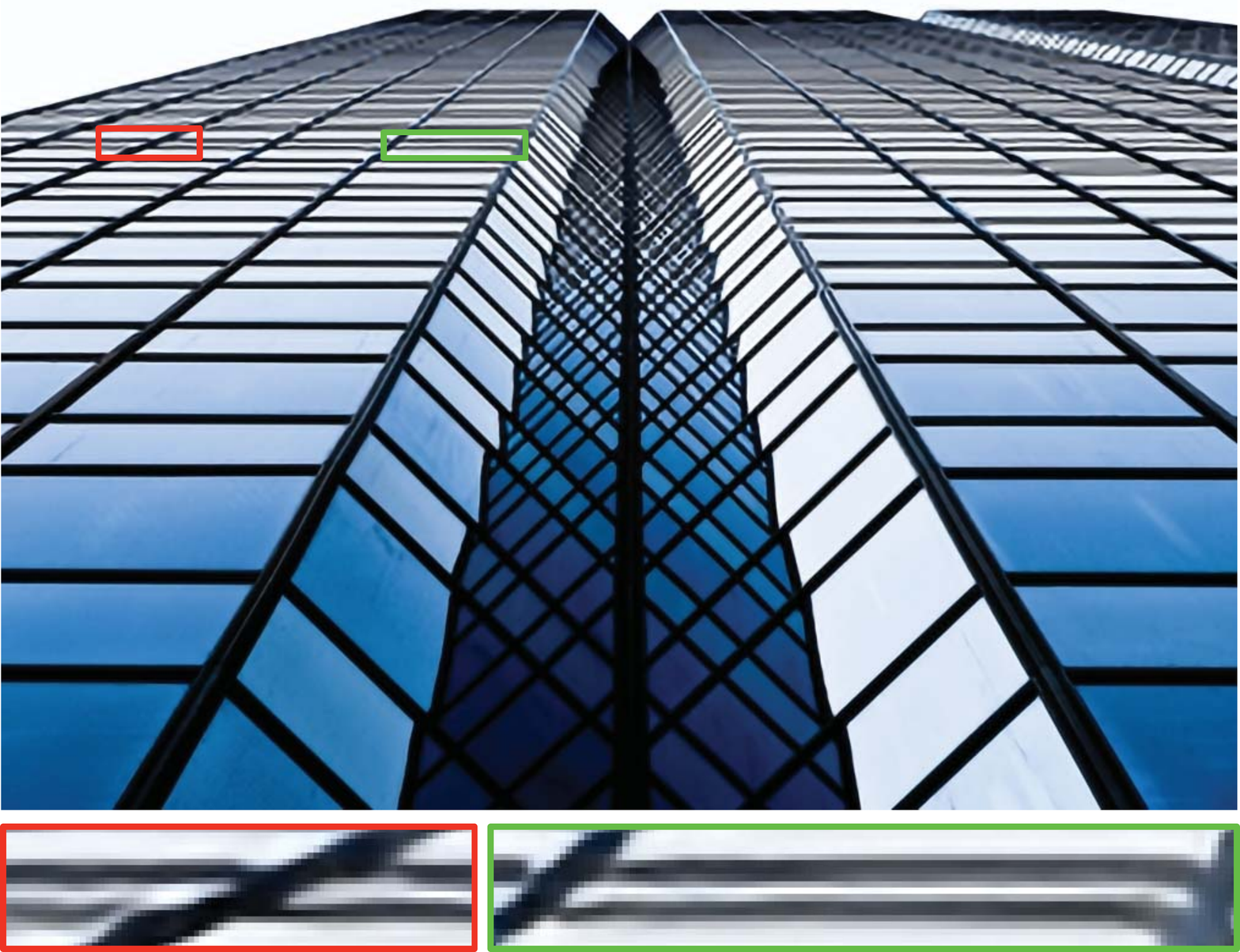}\\
\hspace{-2mm}(a) GT & \hspace{-0.4cm} (b) SRCNN~\cite{SRCNN_eccv14} & \hspace{-0.4cm} (c) VDSR with nearest & \hspace{-0.4cm} (d) VDSR with bilinear & \hspace{-0.4cm} (e) VDSR with bicubic & \hspace{-0.4cm} (f) Ours\\
\end{tabular}
\end{center}
\vspace{-3mm}
\caption{Visual comparisons of super-resolution results by the VDSR method~\cite{VDSR_cvpr16} ($\times 4$) with  structures recovered by different methods, \ie, nearest neighbor, bilinear, and bicubic upsampling.
Residual learning algorithms usually take upsampled image as the base structures and
learn the details, the difference between the upsampled and ground truth images.
However,
residual learning cannot correct low-frequency errors in the structures, \eg, the structure obtained by  nearest neighbor interpolation in (c).
In contrast, our algorithm is motivated by the decomposition of a signal into structures and details, which involves both structure and detail learning and thus leads to better results.
}
\label{fig: sr-structures}
\end{minipage}
\end{figure}

\begin{abstract}
\vspace{-3mm}
In this paper, we propose a general dual convolutional neural network (DualCNN) for low-level vision problems,
e.g., super-resolution, edge-preserving filtering, deraining and dehazing.
These problems usually involve the estimation of two components of the target signals: structures and details.
%
Motivated by this, our proposed DualCNN consists of two parallel branches, which respectively recovers the structures and details in an end-to-end manner.
The recovered structures and details can generate the target signals according to the formation model for each particular application.
The DualCNN is a flexible framework for low-level vision tasks and can be easily incorporated into existing CNNs.
Experimental results show that the DualCNN can be effectively applied to numerous low-level vision tasks with favorable performance against the state-of-the-art methods.
\end{abstract}

\vspace{-4mm}
\section{Introduction}
\vspace{-1mm}
Motivated by the success of deep learning in high-level vision tasks~\cite{Alexnet_nips12,RCNN_fast_iccv15,ResNet_CVPR16,deep_face_application1}, numerous
deep models have been developed for low-level vision tasks, \eg, image super-resolution~\cite{SRCNN_pami16,fast_SRCNN_eccv16,SRCNN_eccv14,VDSR_cvpr16,Recurrent_SR_cvpr16,videoSR_iccv15}, inpainting~\cite{sijieren_sr_nips15,rnnfilter_eccv16},
noise removal~\cite{dong_artifacts_iccv15,denoising_deep_learning_nips08,denoising_deep_learning_nips12}, image filtering~\cite{deepfilter_icml15,rnnfilter_eccv16}, image deraining~\cite{Eigen_derain,derain_gan}, and dehazing~\cite{dehaze_eccv16,DehazeNet_tip16}. 
Although achieving impressive performance, the network architectures of these models strongly resemble those developed for high-level classification tasks.


%
%

Existing methods are based on either plain neural networks or residual learning networks.
As demonstrated in~\cite{Burger-CVPR12, sijieren_sr_nips15}, plain neural networks cannot outperform state-of-the-art traditional approaches on a number of low-level vision problems, e.g., super-resolution~\cite{A+_accv15}.
%
Low-level vision tasks usually involve the estimation of two components, low-frequency structures and high-frequency details. It is challenging for a single network to learn both components simultaneously.
As a result, going deeper with plain neural networks does not always lead to better performance~\cite{SRCNN_pami16}.

Residual learning has been shown to be an effective approach to achieve performance gain with a deeper network.
The residual learning algorithms (e.g.,~\cite{VDSR_cvpr16}) assume that the main structure is given and mainly focus on estimating the residual (details) using a deep network.
These methods work well when the recovered structures are perfect or near perfect.
However, when the main structure is not well recovered, these methods do not perform well, because the final result is a combination of the structures
and details.
Figure~\ref{fig: sr-structures} shows the image super-resolution results by the VDSR method~\cite{VDSR_cvpr16} with  structures recovered by different methods. The residual network cannot correct low-frequency errors in the structures (Figure~\ref{fig: sr-structures}(b)).
To address this issue, we propose a dual convolutional neural network (DualCNN) that can jointly estimate the structures and details.
A DualCNN consists of two branches, one shallow sub-network to estimate the structures and one deep sub-network to estimate the details.
The modular design of a DualCNN makes it a flexible framework for a variety of low-level vision problems.
When trained end-to-end, DualCNNs perform favorably against state-of-the-art methods that have been specially designed for each individual task.



\vspace{-2mm}
\section{Related Work}
\vspace{-2mm}
Numerous deep learning methods have been developed for low-level vision tasks.
A comprehensive review is beyond the scope of this work and we discuss the most related ones in this section.

\vspace{-2mm}
{\flushleft \bf{Super-resolution.}}
The SRCNN~\cite{SRCNN_eccv14} method uses a three-layer plain convolutional neural
network (CNN) for super-resolution.
As the SRCNN method is less effective in recovering image details,
Kim et al.~\cite{VDSR_cvpr16} propose the residual learning~\cite{VDSR_cvpr16} algorithm based on a deeper network.
%
The VDSR algorithm uses the bicubic interpolation of the low-resolution input as the structure
of the high-resolution image and estimates the residual details using a 20-layer CNN.
However, if the image structure is not well recovered, the generated result is likely to contain substantial artifacts, as shown in Figure~\ref{fig: sr-structures}.

\vspace{-2mm}
{\flushleft \bf{Noise/artifacts removal.}} Numerous algorithms based on CNNs have been developed to remove
noise/artifacts~\cite{dong_artifacts_iccv15,denoising_deep_learning_nips08,denoising_deep_learning_nips12} and unwanted components, e.g., rainy/dirty pixels~\cite{Eigen_derain,derain_gan}.
These methods are based on plain models, residual learning models or recurrent models.
In addition, these methods estimate either the output using one plain network, or details
using a residual network.
However, plain networks cannot recover fine details~\cite{ResNet_CVPR16,VDSR_cvpr16} and residual networks cannot correct structural errors.
%
%

\vspace{-2mm}
{\flushleft \bf{Edge-preserving filtering.}}
For edge-preserving filtering, Xu et al.~\cite{deepfilter_icml15} develop a CNN model to approximate a number of filters.
Liu et al.~\cite{rnnfilter_eccv16} use a hybrid network to approximate a number of
edge-preserving filters.
These methods aim to preserve the main structures and remove details using a single network, but this imposes a difficult learning task.
In this work, we show that it is critical to accurately estimate both the structures and the details for low-level vision tasks.

\vspace{-2mm}
{\flushleft \bf{Image dehazing.}}
In image dehazing, existing CNN-based methods~\cite{DehazeNet_tip16,dehaze_eccv16} mainly focus on estimating the transmission map from an input.
Given an estimated transmission map, the atmospheric light can be computed using the air light model.
As such, errors in the transmission maps are propagated into the light estimation process.
For more accurate results, it is necessary to jointly estimate the transmission map and atmospheric light in one model, which DualCNNs are designed for.

A common theme is that we need to design a new network for every low-level vision task.
In this paper, we show that low-level vision problems usually involve the estimation of two components: structures and details.
Thus we develop a single framework, called DualCNN, that can be flexibly applied to a variety of low-level vision problems, including the four tasks discussed above.

\begin{figure*}[!t]\footnotesize
\begin{center}
\begin{tabular}{cc}
\includegraphics[width = 0.86\linewidth]{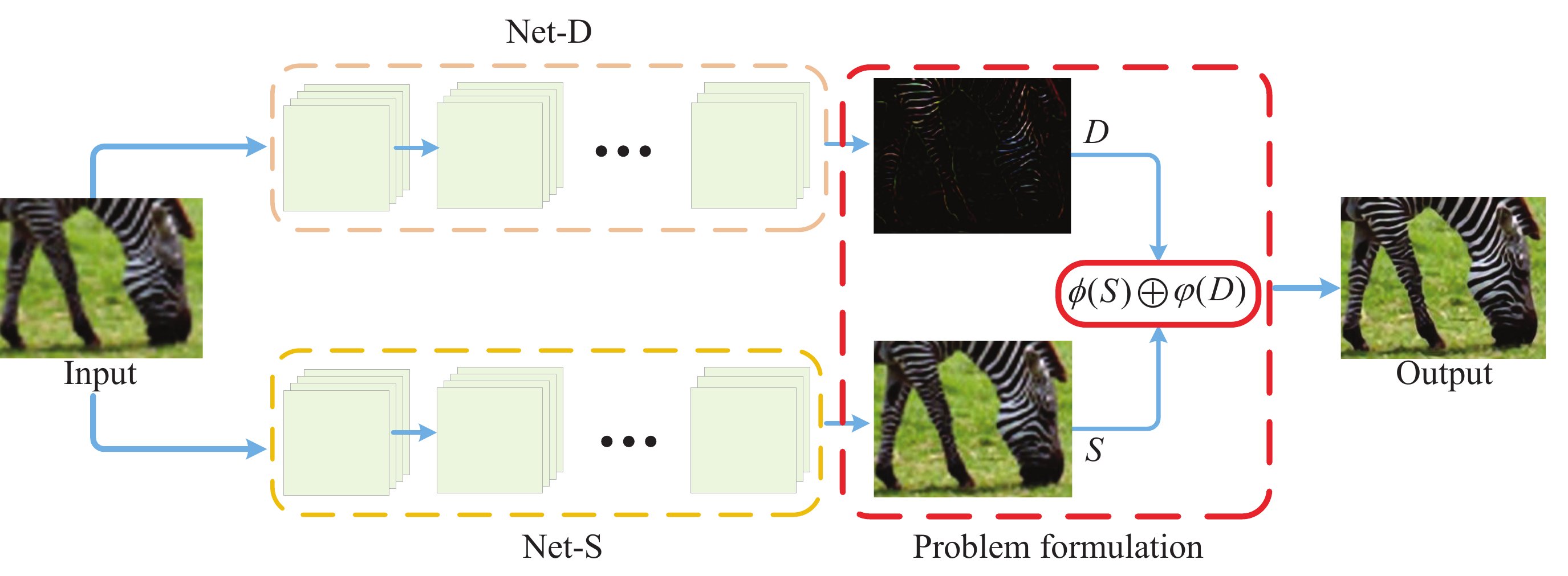}
\end{tabular}
\end{center}
\vspace{-4mm}
\caption{Proposed DualCNN model. It contains two branches, Net-D and Net-S, and a problem formulation module.
A DualCNN first estimates the structures and the details and then reconstructs the final results according to the formulation module.
The whole network is end-to-end trainable.
}
\label{fig: proposed-model}
\end{figure*}

\vspace{-2mm}
\section{Proposed Algorithm}
\vspace{-2mm}
%

As shown in Figure~\ref{fig: proposed-model}, the proposed dual model consists of two branches,  Net-S, and output of Net-D, which respectively estimate the structure and detail components of the target signals from the input.
%
Take image super-resolution as an example.
Given a low-resolution image, we first use the bicubic upsampled image as the input.
Then, our dual network learns the details and structures according to the formulation model of the image decomposition.

\vspace{-3mm}
{\flushleft \bf{Dual composition loss function.}}
Let $X$, $S$, and $D$ denote the ground truth label,  output of Net-S, and output of Net-D, respectively.
The dual composition loss function enforces that the recovered structure $S$ and detail $D$ can generate the ground truth label $X$ using the given formation model:
\vspace{-1mm}
\begin{equation}
\mathcal{L}_x(S, D)  = \|\phi(S) + \varphi(D) - X\|_2^2,
\label{eq: loss-function-whole}
\vspace{-1mm}
\end{equation}
where the forms of the functions $\phi(\cdot)$ and $\varphi(\cdot)$ are known and depend on the domain knowledge of each task.
For example, the functions $\phi(\cdot)$ and $\varphi(\cdot)$ are identity functions for image decomposition problems (e.g., filtering) and restoration problems (e.g., super-resolution, denoising, and deraining).
We will show that $\phi(\cdot)$ and $\varphi(\cdot)$ can take more general forms to deal with specific problems.
%
%

\vspace{-1mm}
\subsection{Regularization of the DualCNN Model}
\vspace{-1mm}

The proposed DualCNN model has two branches, which may cause instability if only the composition loss~\eqref{eq: loss-function-whole} is used.
For example, if Net-S and Net-D have the same structure, symmetrical solutions exist.
To obtain a stable solution, we use individual loss functions to regularize the two branches respectively.
The loss functions for the Net-S and Net-D are defined as
\begin{align}
\mathcal{L}_s(S) & = \|S - S_{gt}\|_2^2,
\label{eq: loss-function-s} \\
\mathcal{L}_d(D) & = \|D - D_{gt}\|_2^2,
\label{eq: loss-function-t}
\end{align}
where $S_{gt}$ and $D_{gt}$ are ground truths corresponding to the outputs of Net-S and
Net-D.
%
%
Consequently the overall loss function to train DualCNN is
\vspace{-1mm}
\begin{equation}
\label{eq: loss-function-final}
\mathcal{L}  = \alpha\mathcal{L}_x + \lambda \mathcal{L}_s + \gamma\mathcal{L}_d,
\vspace{-1mm}
\end{equation}
where $\alpha$, $\lambda$ and $\gamma$ are non-negative trade-off weights.
%
Our framework can also use other loss functions, e.g., perceptual loss for style transfer.

We use the SGD method to minimize  the loss function~\eqref{eq: loss-function-final} and train a DualCNN.
In the training stage, the gradients for Net-S and Net-D can be obtained by
\vspace{-1mm}
\begin{subequations}
\label{eq: back-error}
\begin{align}
&\frac{\partial\mathcal{L}}{\partial S} = 2\alpha\phi'(S)E+ 2\lambda(S - S_{gt}), \label{eq:net-D-subproblem}\\
&\frac{\partial\mathcal{L}}{\partial D} = 2\alpha\varphi'(D)E+ 2\gamma(D - D_{gt}), \label{eq:net-S-subproblem}
\end{align}
\end{subequations}
where $E = \phi(S) + \varphi(D) - X$, $\phi'(S)$ and $\varphi'(D)$ are the derivatives with respect to $S$ and $D$.

In the test stage, we compute the high-quality output $X_{est}$ using the outputs of Net-S and Net-D according to the formation model,
\vspace{-1mm}
\begin{equation}
\label{eq: image-resconstruction}
X_{est}  = \phi(S) + \varphi(D).
\vspace{-1mm}
\end{equation}

%

\vspace{-2mm}
\subsection{Generalization}
\vspace{-2mm}

\label{ssec: generalization}
Aside from image decomposition and restoration problems, the proposed model can handle other low-level vision problems by modifying the composition loss function~\eqref{eq: loss-function-whole}.
Here we use image dehazing as an example.

\vspace{-2mm}
{\flushleft \bf{Image dehazing.}}
The image dehazing model can be described using the air light model,
\vspace{-2mm}
\begin{equation}
\label{eq: haze-process}
I  = JD + S(1-D),
\vspace{-2mm}
\end{equation}
where $I$ is the hazy image, $J$ is the haze-free image, $S$ is the atmospheric light, and $D$ is the medium transmission map,
which describes the portion of the light that reaches the camera from scene surfaces.
With the formulation model~\eqref{eq: haze-process}, we can set $\phi(S) = S(1-D)$ and $\varphi(D) = JD$ in~\eqref{eq: loss-function-whole} within the DualCNN framework.
As a result, the composition loss function~\eqref{eq: loss-function-whole} for image dehazing becomes
\vspace{-2mm}
\begin{equation}
\label{eq: loss-function-whole-hazy}
\mathcal{L}_x(S, D)  = \|JD + S(1-D) - I\|_2^2.
\vspace{-2mm}
\end{equation}
The other two loss functions~\eqref{eq: loss-function-s} and \eqref{eq: loss-function-t} remain the same.
In the training phase, we use the same method~\cite{dehaze_eccv16} to generate the atmospheric light $S$,  the transmission map $D$ and construct hazy/haze-free image pairs.
The implementation details of the training stage are presented in Section~\ref{ssec: image-dehazing}.

In the test phase, the clear image $J_{est}$ can be reconstructed by the outputs of Net-D and Net-S, i.e.,
\vspace{-2mm}
\begin{equation}
\label{eq: haze-image-estimation}
J_{est}  = \frac{I - S}{\max\{D, d_0\}} + S,
\vspace{-2mm}
\end{equation}
where  $d_0$ is used to prevent division by zero and a typical value is $0.1$ .

\vspace{-1mm}
\section{Experimental Results}
\vspace{-2mm}
\label{sec: Experimental Results}
We evaluate DualCNNs on several low-level vision tasks including super-resolution, edge-preserving smoothing, deraining and dehazing.
The main results are presented in this section and more results can be found in the supplementary material.
The trained models are publicly available on the authors' websites.
\vspace{-3mm}
{\flushleft \bf{Network parameters.}}
Motivated by the success of SRCNN and VDSR for super-resolution, we use 3 convolution layers followed by the ReLU function for the network Net-S.
The filter sizes of each layer are $9\times 9$, $1\times 1$, and $5\times 5$, respectively.
The depths of each layer are $64$, $32$, and $1$, respectively.
For the network Net-D, we use 20 convolution layers followed by the ReLU function.
The filter size of each layer is $3\times 3$ and the depth of each layer is $64$.
The batch size is set to be $64$ and the learning rate is $10^{-4}$.
Although each branch of the proposed model is similar to SRCNN or VDSR, both our analysis and experimental results show that the proposed model is significantly different from these methods and achieves better results.

\begin{figure*}[!t]\footnotesize
\begin{center}
\begin{tabular}{cccc}
\includegraphics[width = 0.22\linewidth, height = 0.29\linewidth]{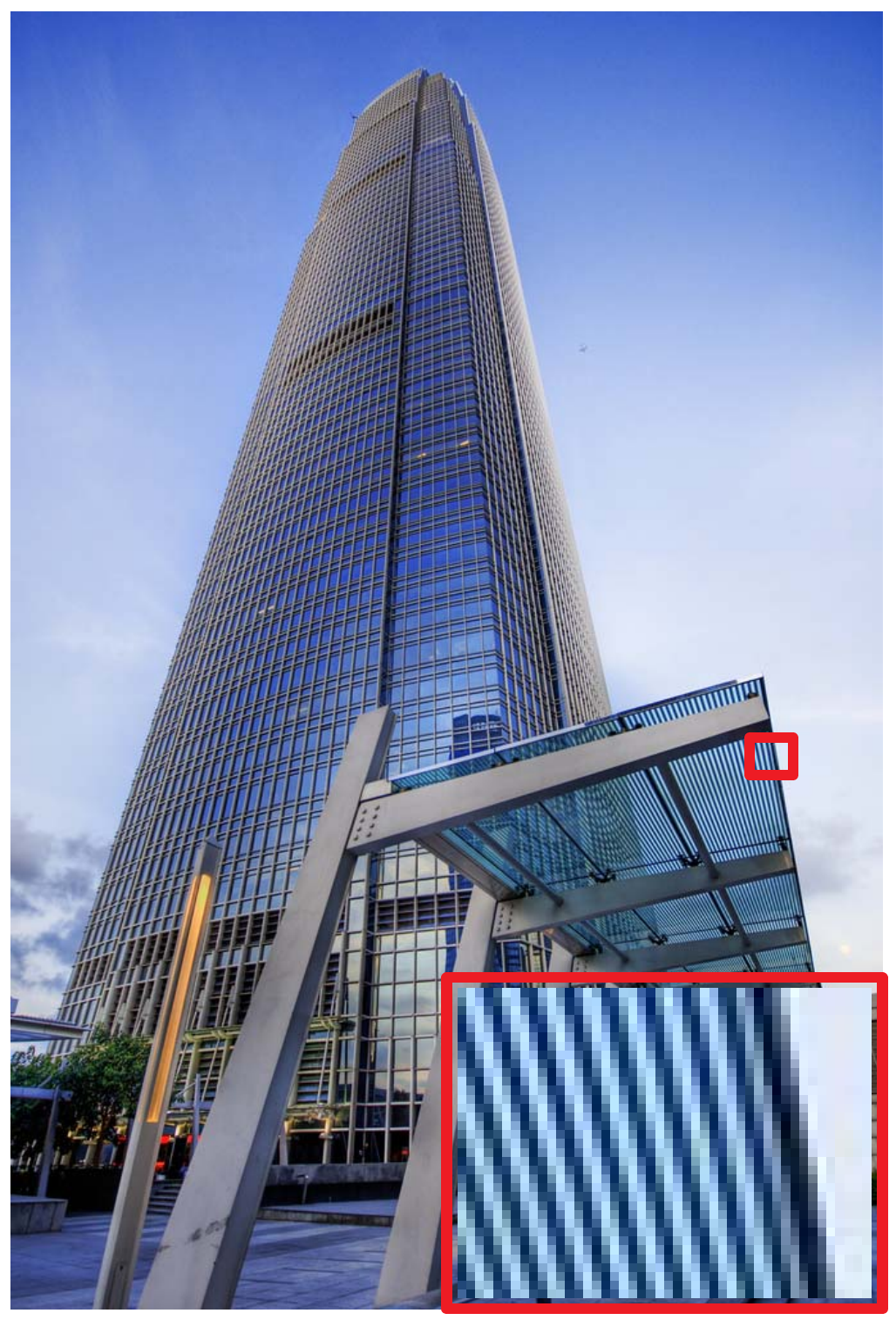}& \hspace{-4mm}
\includegraphics[width = 0.22\linewidth, height = 0.29\linewidth]{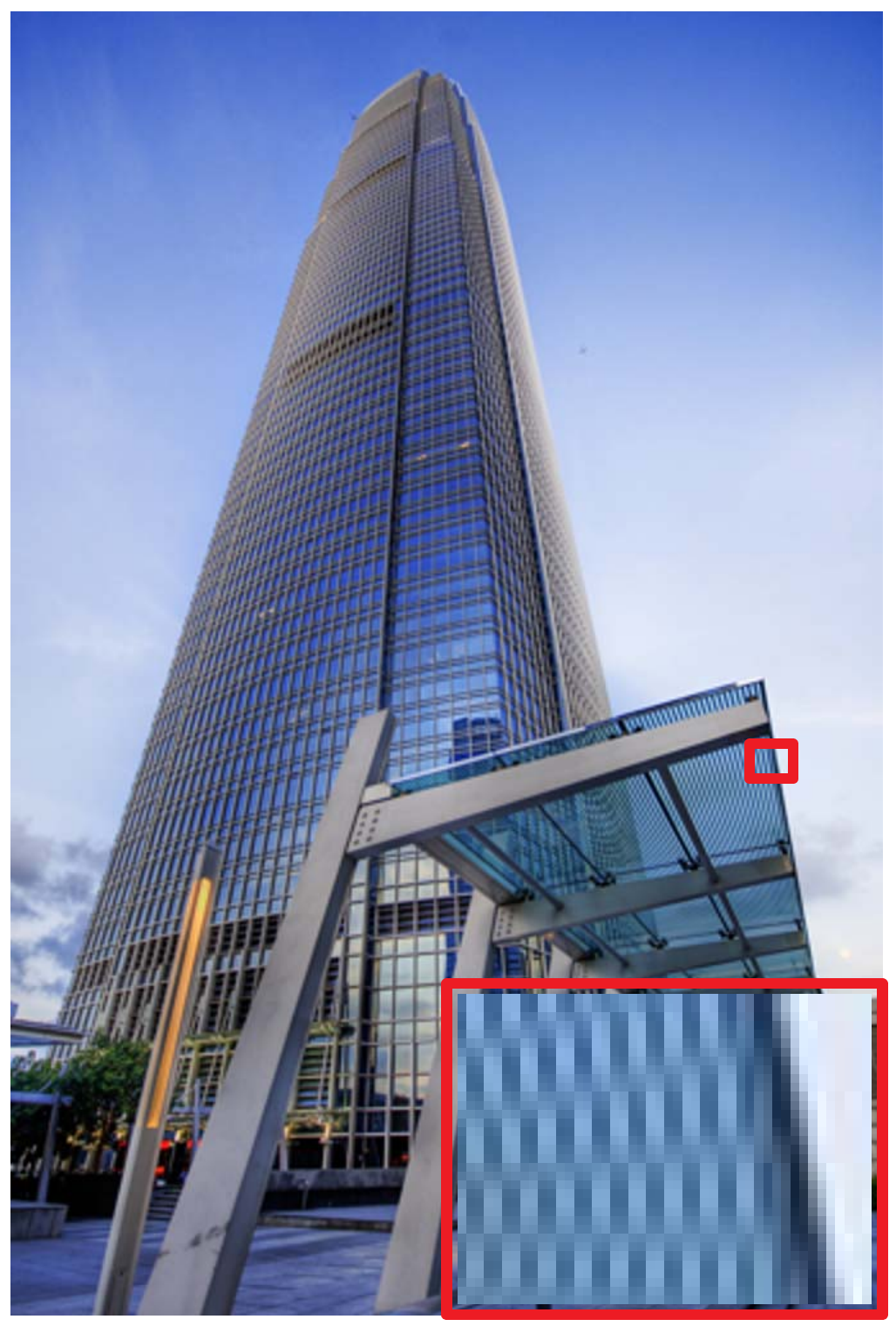}& \hspace{-4mm}
\includegraphics[width = 0.22\linewidth, height = 0.29\linewidth]{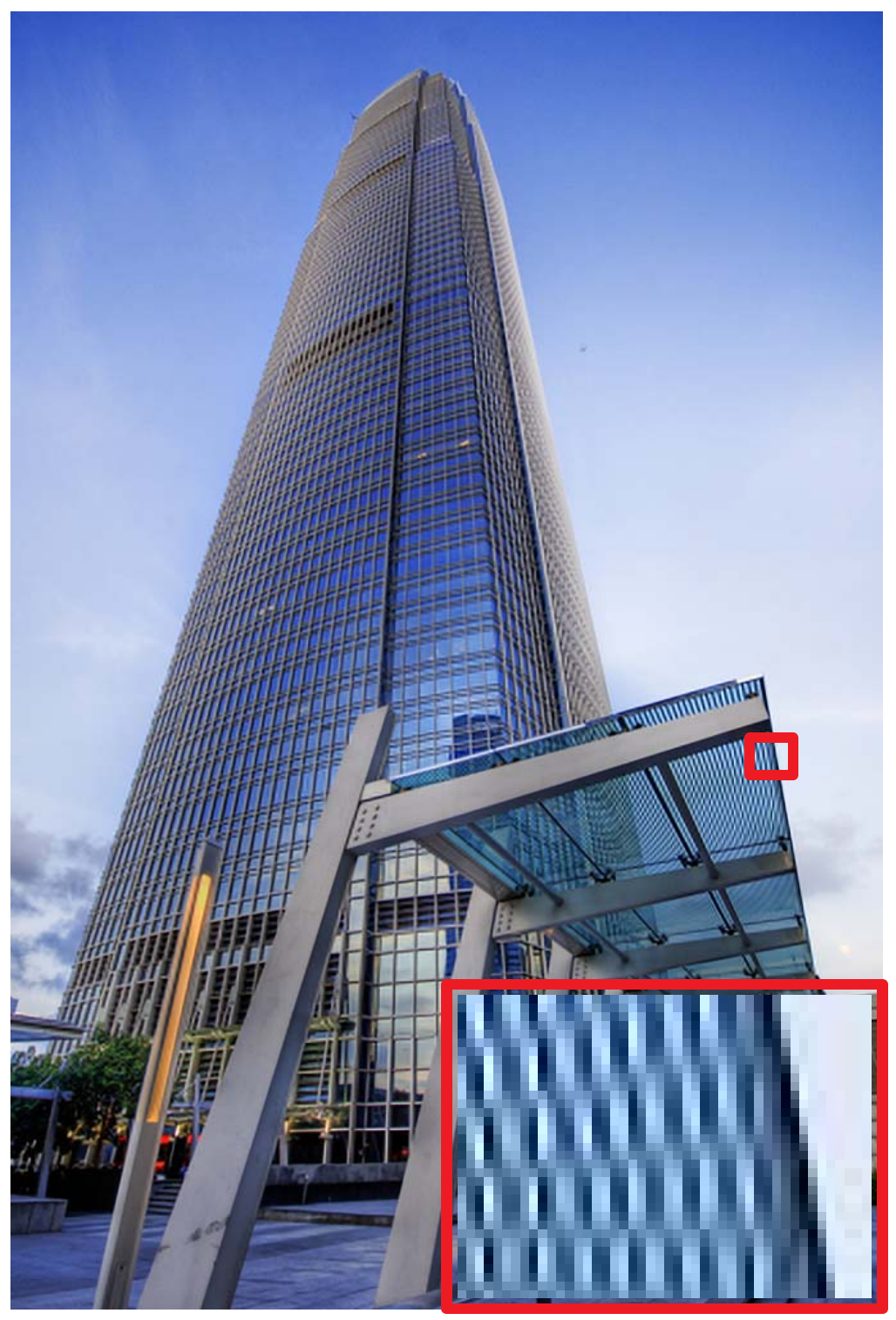} &\hspace{-4mm}
\includegraphics[width = 0.22\linewidth, height = 0.29\linewidth]{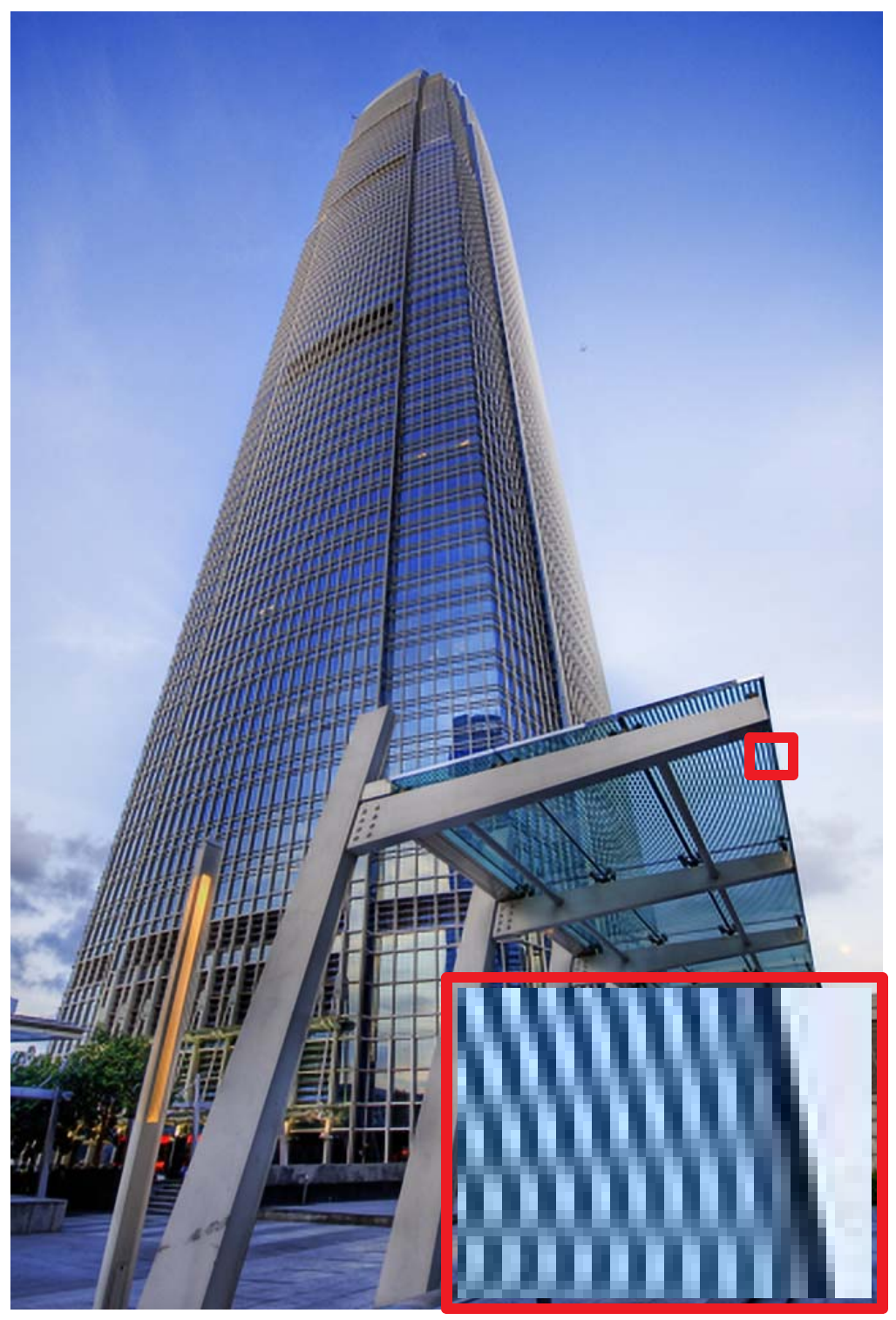}\\
(a) GT &\hspace{-4mm} (b) Bicubic &\hspace{-4mm} (c) SelfEx &\hspace{-4mm} (d) SRCNN\\
\includegraphics[width = 0.22\linewidth, height = 0.29\linewidth]{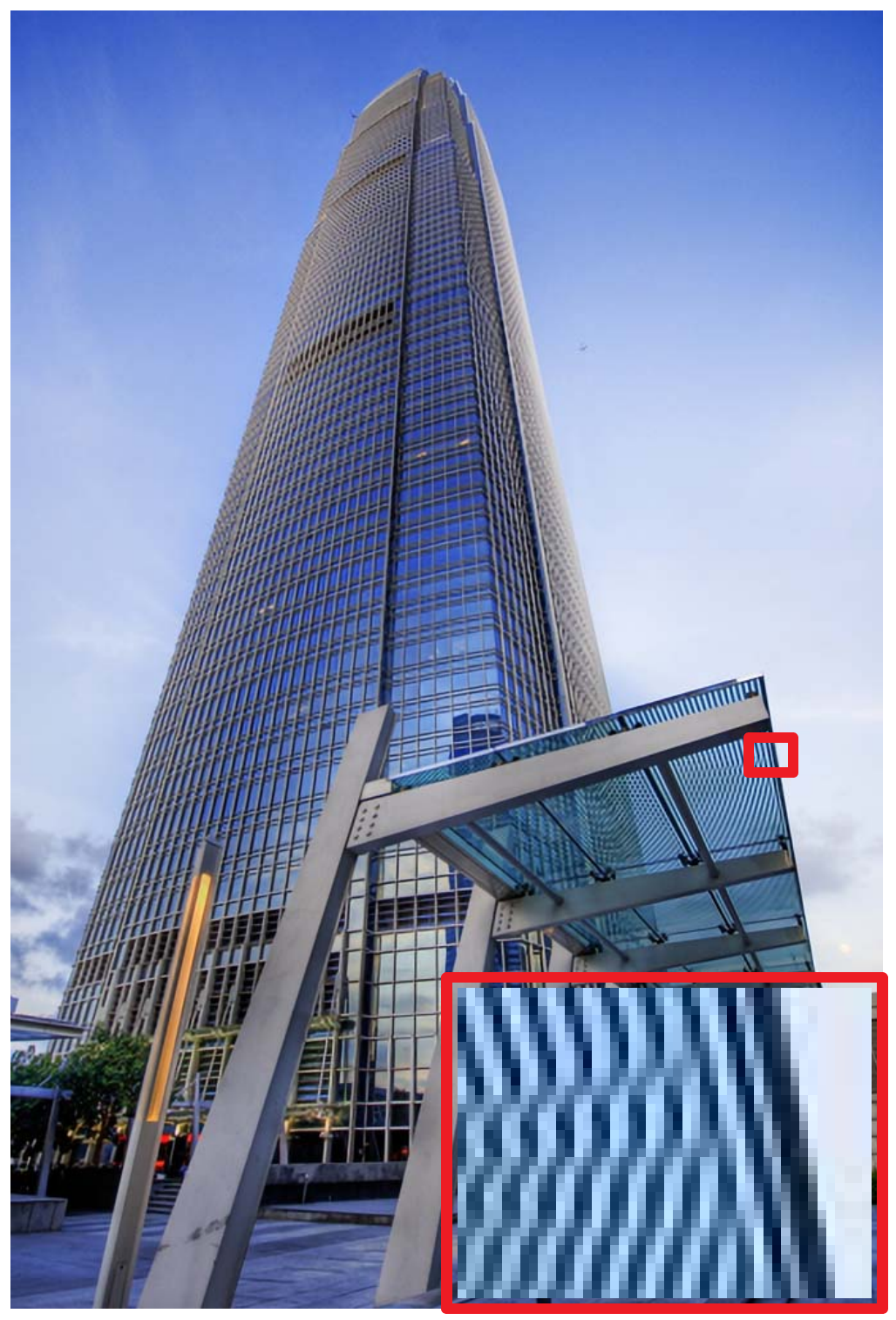} & \hspace{-4mm}
\includegraphics[width = 0.22\linewidth, height = 0.29\linewidth]{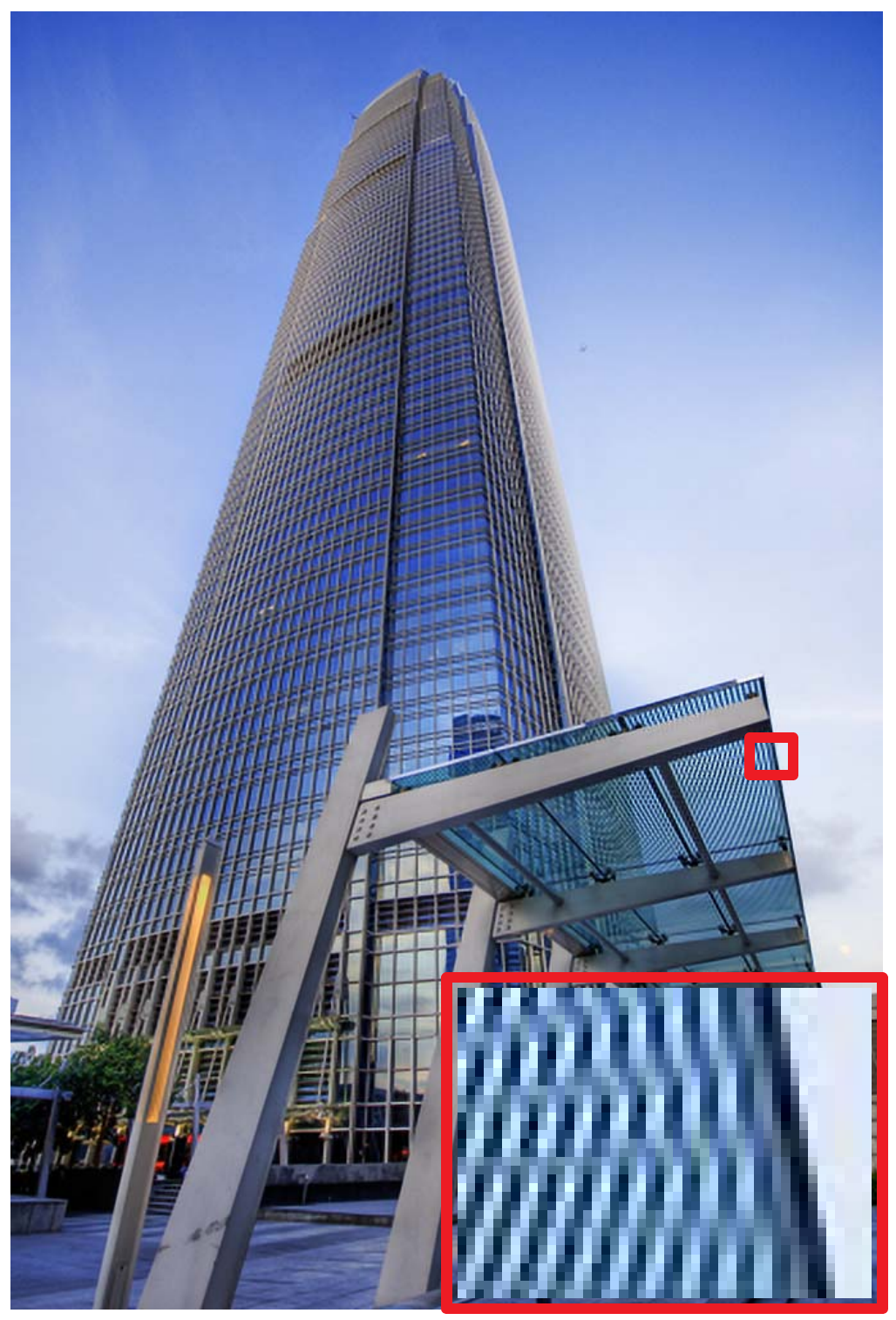} & \hspace{-4mm}
\includegraphics[width = 0.22\linewidth, height = 0.29\linewidth]{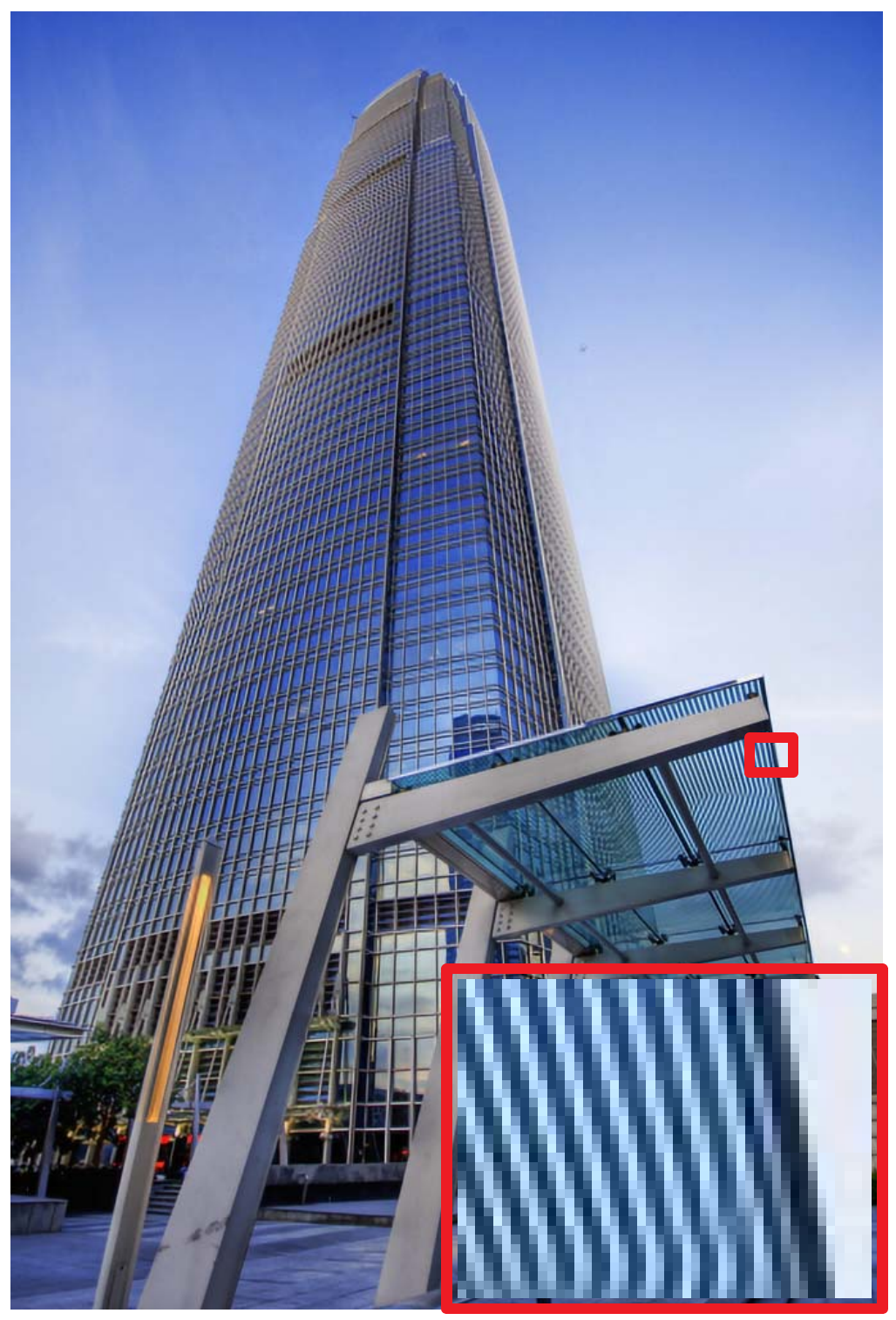}& \hspace{-4mm}
\includegraphics[width = 0.22\linewidth, height = 0.29\linewidth]{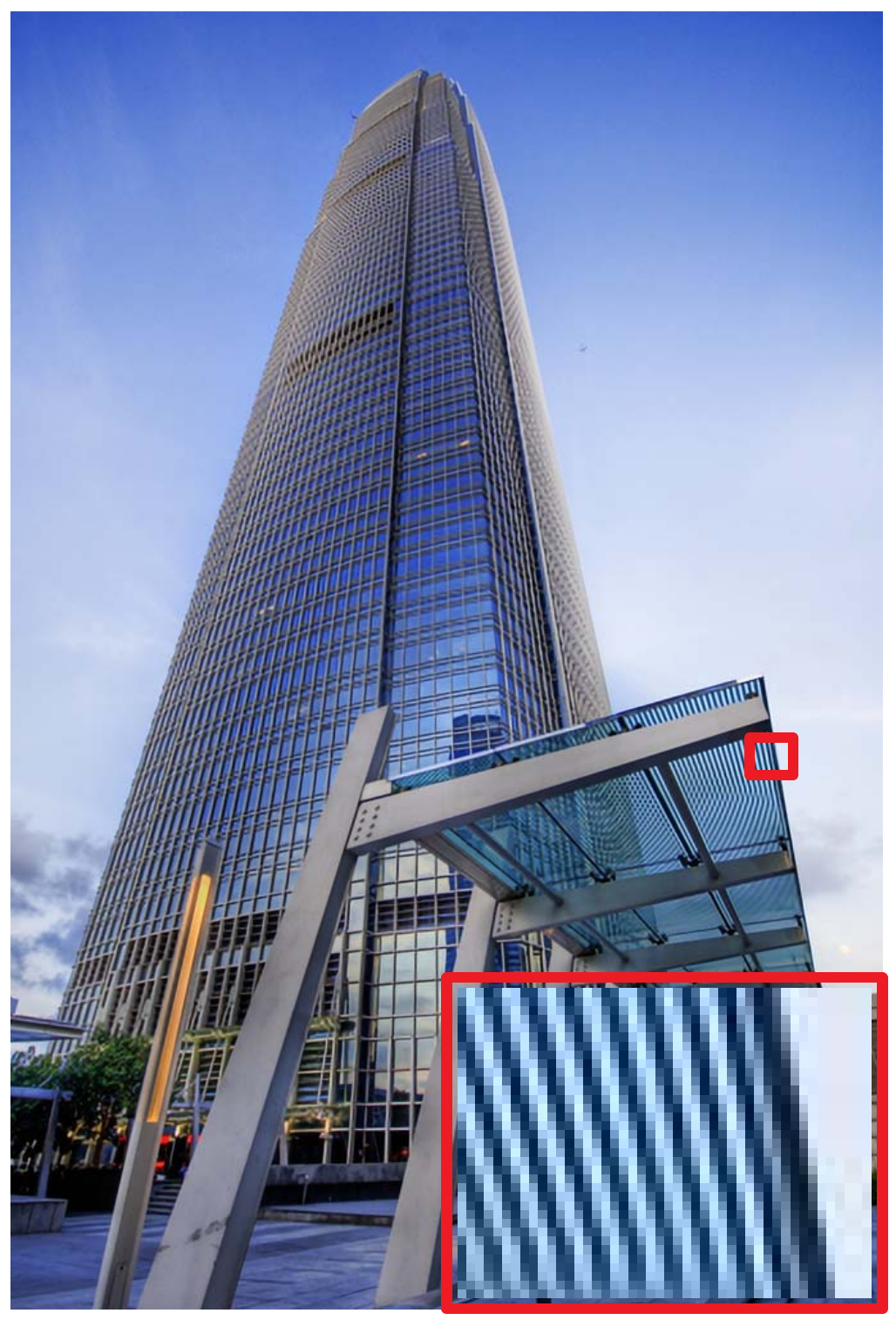}\\
 (e) VDSR &\hspace{-4mm} (f) ESPCN &\hspace{-4mm} (g) SRGAN &\hspace{-4mm} (h) Ours\\
\end{tabular}
\end{center}
\vspace{-3mm}
\caption{Visual comparisons for super-resolution ($\times 2$).
The state-of-the-art methods do not preserve the main structures of the images, while the proposed method generates a better result.}
\label{fig: sr-visualization}
\end{figure*}

\begin{table*}[!t]
\footnotesize
  \caption{Quantitative evaluations for the state-of-the-art super-resolution methods on the benchmark datasets (Set5, Set14, Urban100, and BSDS500) in terms of PSNR and SSIM.
  }
   \vspace{1mm}
   \label{tab:psnr-sr}
 \resizebox{\textwidth}{!}{
 \centering
 \begin{tabular}{ccccccccccc}
    \toprule
    \multirow{2}{*}{Dataset} &\multirow{2}{*}{Scale}&Bicubic&A+&SelfEx&SRCNN&ESPCN&VDSR&SRGAN&Ours\\
                                                  & &PSNR/SSIM&PSNR/SSIM&PSNR/SSIM&PSNR/SSIM&PSNR/SSIM&PSNR/SSIM&PSNR/SSIM&PSNR/SSIM\\
    \hline
    \multirow{3}{*}{Set5} &$\times 2$&33.6924/0.9308&36.575/0.9546&36.5392/0.9537&36.4191/0.9531&36.7315/0.9547&37.6173/0.9596&37.0098/0.9548&\bf{37.7005}/\bf{0.9600}\\
                          &$\times 3$&30.4396/0.8694&32.6866/0.9097&32.6759/0.9099&32.4957/0.9049&32.6880/0.9077&33.7571/0.9229&33.5384/0.9170&\bf{33.8003}/\bf{0.9234}\\
                          &$\times 4$&28.4528/0.8116&30.3471/0.8623&30.3458/0.8636&30.1496/0.8551&30.2730/0.8540&31.4657/\bf{0.8863}&31.3496/0.8797&{\bf{31.4778}}/0.8860\\
    \hline
     \multirow{3}{*}{Set14}&$\times 2$&30.2660/0.8687&32.3497/0.9051&32.2657/0.9029&32.2934/0.9040&32.4020/0.9056&33.2037/0.9131&32.6889/0.9049&\bf{33.2334}/\bf{0.9131}\\
     					  &$\times 3$&27.5556/0.7731&29.1602/0.8181&29.1743/0.8190&29.0676/0.8147&29.1161/0.8161&29.8720/\bf{0.8319}&29.5351/0.8227&{\bf{29.8822}}/0.8315\\
     					  &$\times 4$&26.0089/0.7006&27.3238/0.7481&27.3956/0.7509&27.2283/0.7402&27.1681/0.7401&28.0667/\bf{0.7671}&27.8353/0.7588&{\bf{28.0949}}/0.7669\\
    \hline
   \multirow{3}{*}{Urban100}
   						  &$\times 2
   						  $&26.8621/0.8400&28.5485/0.8782&29.5317/0.8962&29.1009/0.8896&29.2381/0.8920&30.7897/0.9144&29.4390/0.8745&\bf{30.8273}/\bf{0.9145}\\
                          &$\times 3 $&24.4375/0.7336&25.7907/0.7878&26.4188/0.8079&25.8549/0.7874&25.9170/0.7897&27.1297/\bf{0.8278}&26.6243/0.8159&{\bf{27.1318}}/0.8277\\
                          &$\times 4 $&23.1158/0.6551&24.1890/0.7119&24.7648/0.7361&24.1275/0.7030&24.1534/0.7031&25.1575/\bf{0.7515}&24.1516/0.7298&{\bf25.1722}/0.7510\\
    \hline
   \multirow{3}{*}{BSDS500} &$\times 2$&29.6393/0.8622&30.8758/0.8929&31.3447/0.9016&31.3319/0.9013&31.4187/0.9030&32.2226/0.9136&31.3938/0.8889&\bf{32.2458}/\bf{0.9140}\\
                          &$\times 3$&27.1875/0.7626&28.1461/0.8024&28.2960/0.8073&28.2233/0.8033&28.2404/0.8048&28.8889/0.8229&28.6354/0.8159&\bf{28.8979}/\bf{0.8232}\\
                          &$\times 4$&25.8953/0.6931&26.6798/0.7324&26.7851/0.7368&26.6595/0.7278&26.6122/0.7278&27.2342/0.7525&26.9104/0.7423&\bf{27.2454}/\bf{0.7527}\\
 \bottomrule
  \end{tabular}
}
\end{table*}

\begin{table*}[!t]\footnotesize
  \caption{Average running time (seconds) of the evaluated methods on the test dataset~\cite{VDSR_cvpr16}.}
   \vspace{1mm}
  \label{tab:run-time}
  \centering
  \begin{tabular}{lccccc}
    \toprule
     Methods &A+&SelfEx& SRCNN & VDSR & Ours\\
    \midrule
    Average running time &0.88&99.04&0.55&4.85 & 5.19\\
    \bottomrule
  \end{tabular}
\end{table*}

\begin{table*}[!t]
\footnotesize
  \caption{PSNR results for learning various image filters on the test dataset~\cite{deep/deconvolution}.
   \vspace{1mm}
  }
  \label{tab:psnr-filtering}
  \centering
  \begin{tabular}{lccccc}
    \toprule
     &Xu et al.~\cite{deepfilter_icml15}&Liu et al.~\cite{rnnfilter_eccv16}& VDSR~\cite{VDSR_cvpr16} &Net-S& Ours\\
    \midrule
    $L_0$ &\bf{32.8}&30.9&31.5& 28.0&31.4\\
    WMF &31.4&34.0&38.5&29.2&\bf{39.1}\\
    RTV &32.1&37.1&41.6&32.0&\bf{42.1}\\
    \bottomrule
  \end{tabular}
\end{table*}

\begin{figure*}[!t]\footnotesize
\begin{center}
\begin{tabular}{cccccc}
\includegraphics[width = 0.16\linewidth, height = 0.21\linewidth]{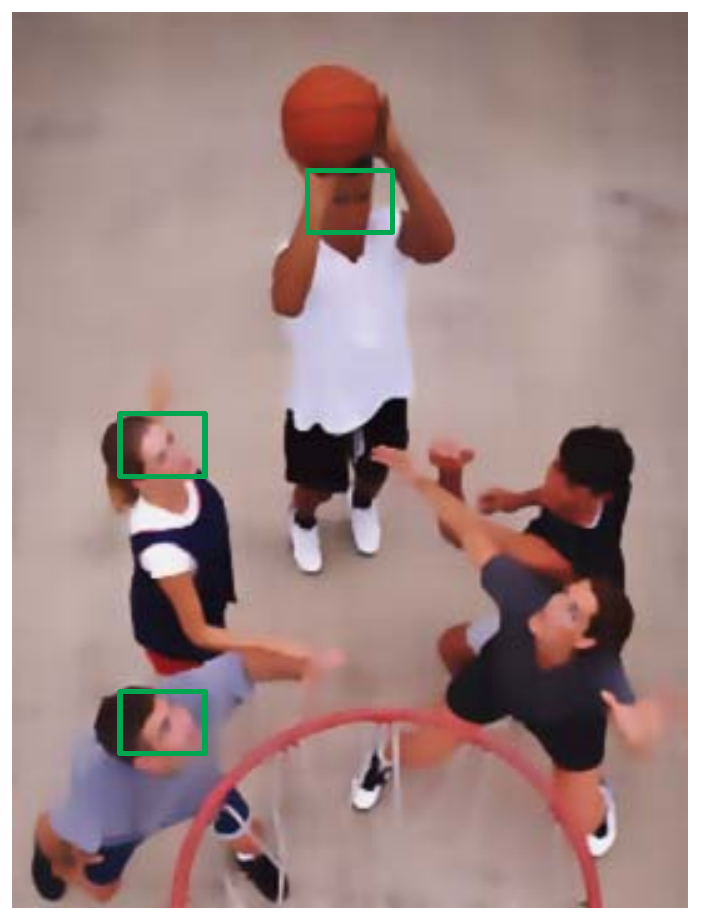}& \hspace{-5mm}
\includegraphics[width = 0.16\linewidth, height = 0.21\linewidth]{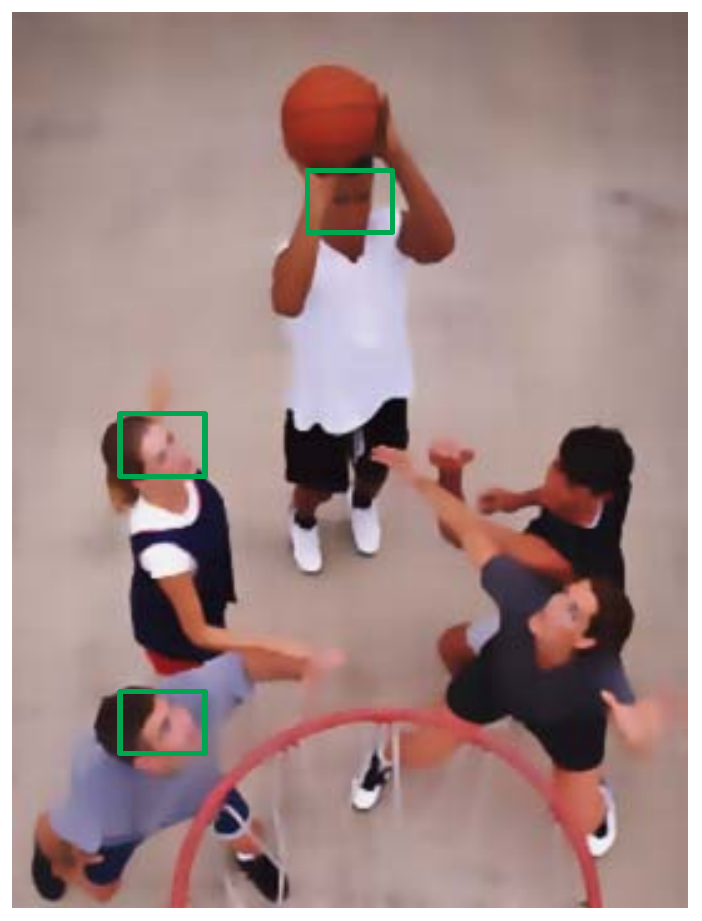}& \hspace{-5mm}
\includegraphics[width = 0.16\linewidth, height = 0.21\linewidth]{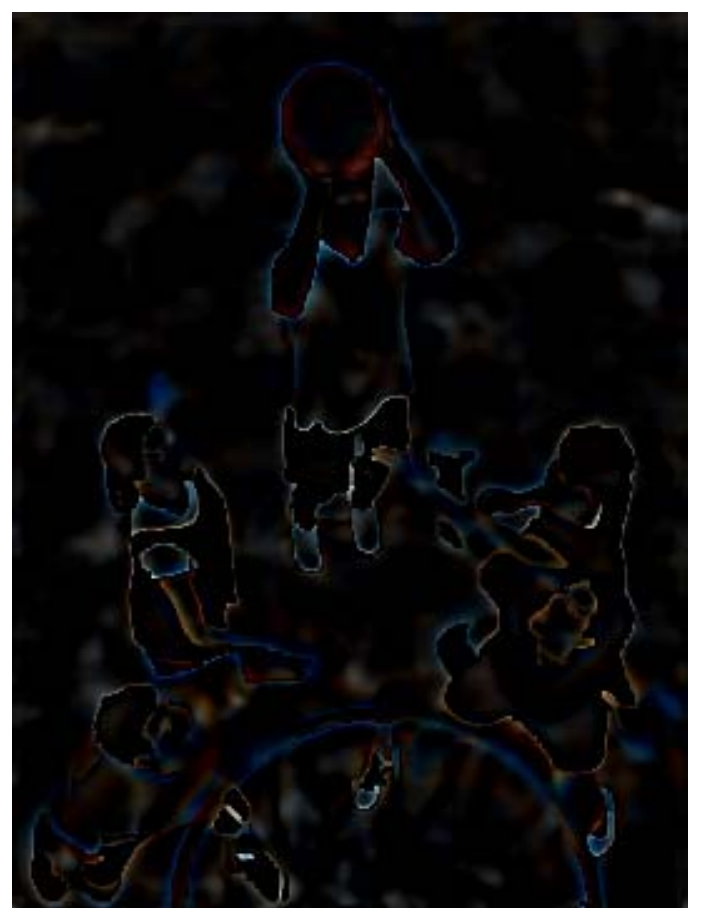}& \hspace{-5mm}
\includegraphics[width = 0.16\linewidth, height = 0.21\linewidth]{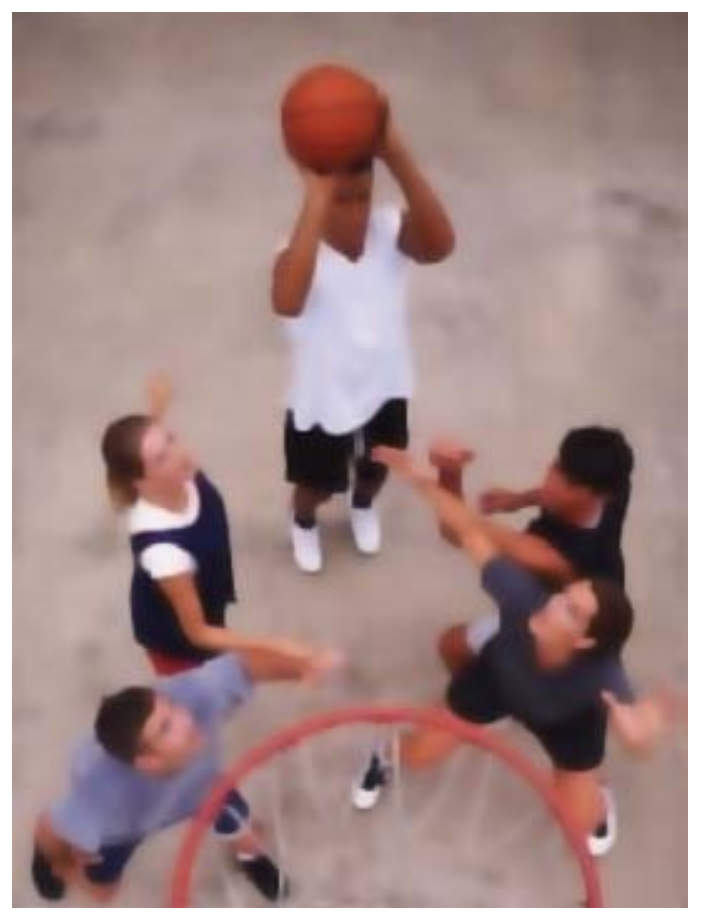} & \hspace{-5mm}
\includegraphics[width = 0.16\linewidth, height = 0.21\linewidth]{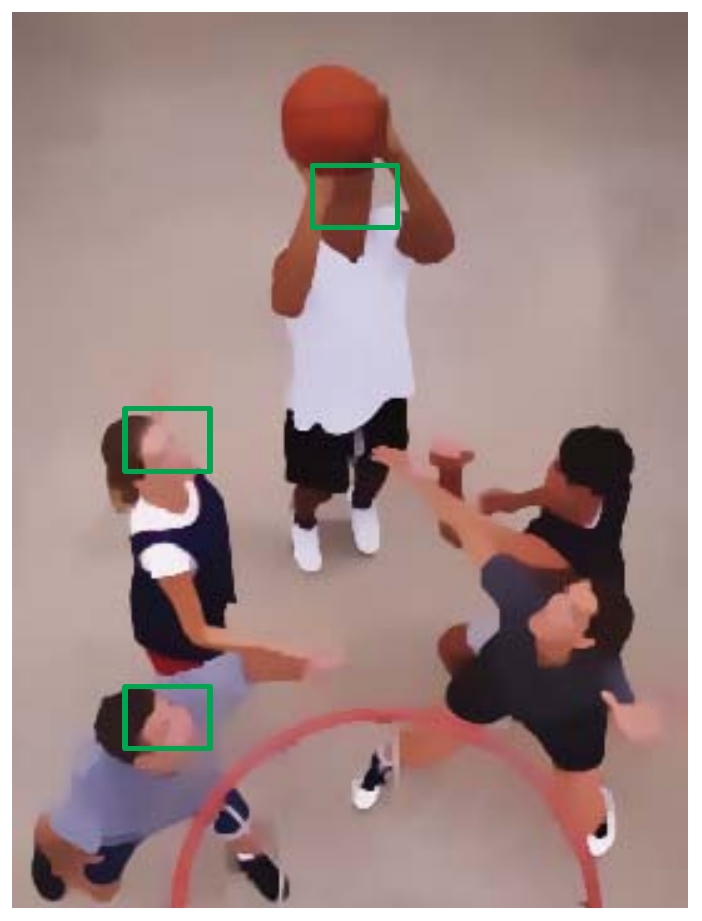}& \hspace{-5mm}
\includegraphics[width = 0.16\linewidth, height = 0.21\linewidth]{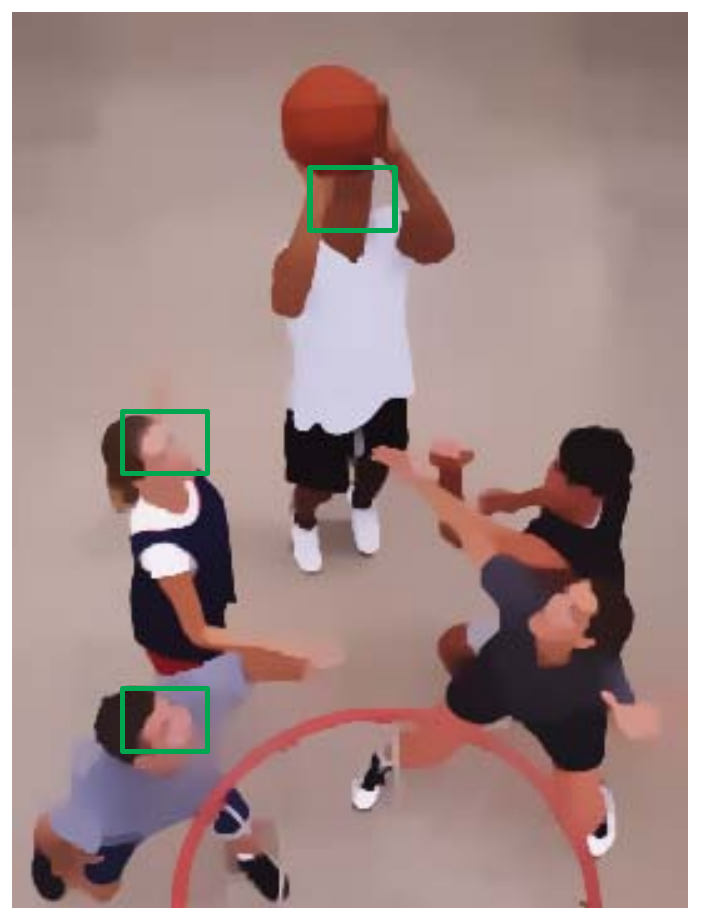}\\
(a) Xu et al. &\hspace{-4mm} (b) Liu et al. &\hspace{-4mm} (c) Net-D &\hspace{-4mm} (d) Net-S &\hspace{-4mm} (e) Ours &\hspace{-4mm} (f) RTV\\
\end{tabular}
\end{center}
\vspace{-3mm}
\caption{Visual comparisons for learning the relative total variation (RTV) image filters.
Existing deep learning based methods are not able to remove the details and structures that are supposed to be removed (the green boxes in (a) and (b)).
(c) and (d) show the outputs of the two branches of the proposed model. (f) is the result by the original implementation of RTV. Better enlarge and view on a screen.
}
\label{fig: rtv-filtering}
\end{figure*}

\vspace{-1mm}
\subsection{Image Super-resolution}
\vspace{-1mm}
\vspace{-1mm}
{\flushleft \bf{Training data.}}
For image super-resolution, we generate the training data by randomly sampling 250 thousands $41\times 41$ patches from 291 natural images in~\cite{BSDS}.
We apply the Gaussian filter to each ground truth label $X$ to obtain $S_{gt}$. The ground truth $D_{gt}$ is the difference between the ground truth label $X$ and the structure $S_{gt}$.

For this application, we set $\phi(S) = S$ and $\varphi(D) = D$. The weights $\alpha$, $\lambda$ and $\gamma$ in the loss function~\eqref{eq: loss-function-final} are set to be $1$, $0.001$ and $0.01$, respectively. To increase the accuracy, we use the pre-trained models of SRCNN and VDSR as the initializations of Net-S and Net-D.


We present quantitative and qualitative comparisons against the
state-of-the-art methods including A+~\cite{A+_accv15}, SelfEx~\cite{SelfEx_sr}, SRCNN~\cite{SRCNN_eccv14}, ESPCN~\cite{ESPCN}, SRGAN~\cite{SRGAN}, and VDSR~\cite{VDSR_cvpr16}.
Table~\ref{tab:psnr-sr} shows quantitative evaluations on benchmark datasets.
Overall, the proposed method performs favorably against the state-of-the-art methods.
Note that the architecture of one branch in a DualCNN is either similar to SRCNN or VDSR.
However, the results generated by a DualCNN have highest average PSNR values,
suggesting the effectiveness of the proposed dual model.
%
Figure~\ref{fig: sr-visualization} shows some super-resolution results by the evaluated methods.
The proposed algorithm can well preserve the main structures than state-of-the-art methods.

\vspace{-3mm}
{\flushleft \bf{Running time.}}
We benchmark the running time of all methods on a machine with an Intel Core i7-7700 CPU and an NVIDIA GTX 1080 GPU.
Table~\ref{tab:run-time} shows that the running time of the DualCNN model is comparable to VDSR, which achieves state-of-the-art results on the super-resolution benchmark dataset~\cite{VDSR_cvpr16}.

\vspace{-1mm}
\subsection{Edge-preserving Filtering}
\vspace{-1mm}
\label{ssec: Edge-Preserving Filtering}
%
%
Similar to the methods in~\cite{deepfilter_icml15} and~\cite{rnnfilter_eccv16}, we apply the DualCNN to learn edge preserving image filters including $L_0$ smoothing~\cite{xul0_smooth_tog11}, relative total variation (RTV)~\cite{rtv_sa12}, and weighted median filter (WMF)~\cite{wmf}.
We generate the training data by randomly sampling 1 million patches (clear/filtered pairs) from 200 natural images in~\cite{BSDS}.
Each image patch is of $64\times 64$ pixels, and other settings of generating training data are the same as those used in~\cite{deepfilter_icml15}.

For this application, as our goal is to learn the filtered image which does not contain rich details,
we set weights $\alpha$, $\lambda$, and $\gamma$ in the loss function~\eqref{eq: loss-function-final} to be $1$, $10^{-4}$ and $0$, respectively.
We further let $S_{gt}$ be the ground truth label $X$.
%

We evaluate the proposed DualCNN model against methods~\cite{deepfilter_icml15,rnnfilter_eccv16} using the dataset from~\cite{deepfilter_icml15}.
Table~\ref{tab:psnr-filtering} summarizes the PSNR results.
Note that Xu et al.~\cite{deepfilter_icml15} use image gradients to train their model and the final results are reconstructed by solving a constrained optimization problem.
Thus it performs better for approximating $L_0$ smoothing.
However, our method does not need these additional steps and generates high quality filtered images with significant improvements over the state-of-the-art deep learning based methods,
particularly on RTV and WMF.

We note that the architecture of Net-D is similar to that of VDSR.
As such, we retrain the network of VDSR for these problems.
The results in Table~\ref{tab:psnr-filtering} suggest that only using residual learning does not always generate high-quality filtered images.

Figure~\ref{fig: rtv-filtering} shows the filtering results of approximating RTV~\cite{rtv_sa12}.
The state-of-the-art methods~\cite{deepfilter_icml15,rnnfilter_eccv16} fail to smooth
the structures (e.g., the eyes in the green boxes) that are supposed to be removed
using the RTV filter (Figure~\ref{fig: rtv-filtering}(f)).
In addition, the results with only one branch (i.e., Net-S) have lower PSNR values (Table~\ref{tab:psnr-filtering}) and some remaining tiny structures (Figure~\ref{fig: rtv-filtering}(d)).
%
In contrast, joint learning structures and details preserves more accurate results and the filtered images are significantly closer to the ground truth.

\vspace{-2mm}
\subsection{Image Deraining}
\label{ssec: image-deraining}
\vspace{-2mm}
Deraining aims to recover clear contents from rainy images.
This process can be regarded as recovering the clear details (rainy streaks) and structures (clear images) from inputs.
%
We evaluate the proposed DualCNN on this task.

To train the proposed DualCNN for image deraining, we generate the training data by randomly sampling 1 million patches (rainly/clear pairs) from the rainy image dataset used in~\cite{derain/gan}.
The size of each image patch used in training stage is $64\times 64$ pixels.
Following settings used in learning image filtering, we let $S_{gt}$ be the ground truth label $X$ (i.e., clear image patch).
The weights $\alpha$, $\lambda$ and $\gamma$ in the loss function~\eqref{eq: loss-function-final} are set to be $1$, $0.01$, and $0$, respectively.
We use the test dataset~\cite{derain_gan} to evaluate the effectiveness of the proposed method.
%

Table~\ref{tab:psnr-derain} shows the average PSNR values of restored images on the test dataset~\cite{derain_gan}.
Overall, the proposed method generates the results with the highest PSNR values.

\begin{figure*}[!t]\footnotesize
\begin{center}
\begin{tabular}{cccccc}
\hspace{-4mm}
\includegraphics[width = 0.16\linewidth]{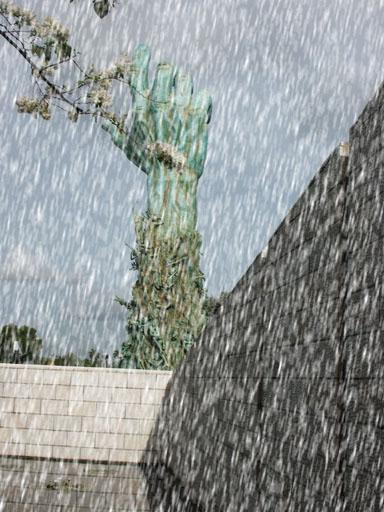}& \hspace{-4mm}
\includegraphics[width = 0.16\linewidth]{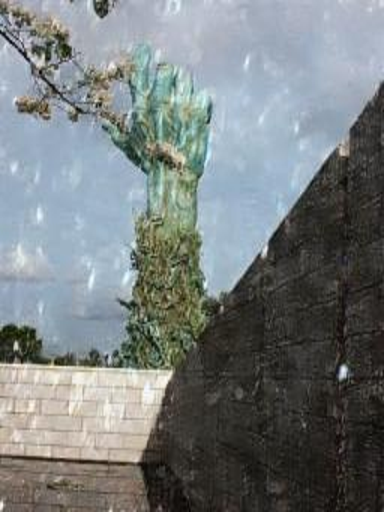}& \hspace{-4mm}
\includegraphics[width = 0.16\linewidth]{figures/derain/69_gan}& \hspace{-4mm}
\includegraphics[width = 0.16\linewidth]{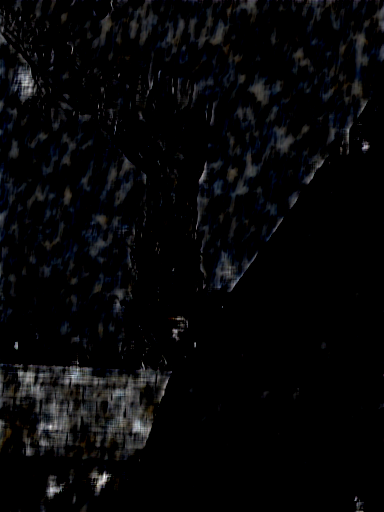} & \hspace{-4mm}
\includegraphics[width = 0.16\linewidth]{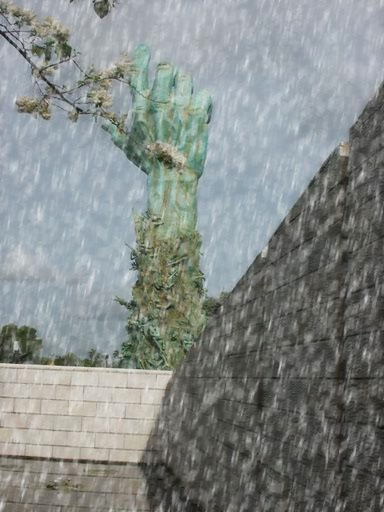}& \hspace{-4mm}
\includegraphics[width = 0.16\linewidth]{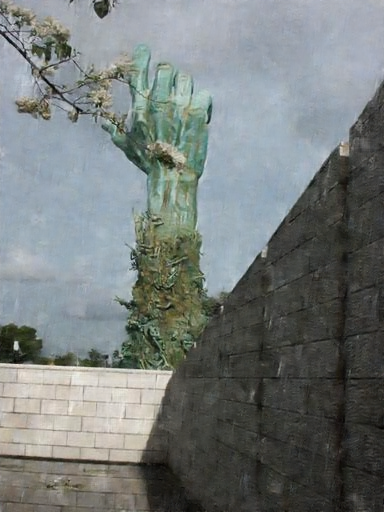}\\
\hspace{-4mm}(a) Input &\hspace{-4mm} (b) SPM &\hspace{-4mm} (c) ID-CGAN &\hspace{-4mm} (d) Net-D &\hspace{-4mm} (e)  Net-S &\hspace{-4mm} (f) Ours \\
\end{tabular}
\end{center}
\vspace{-3mm}
\caption{Visual comparisons for image deraining. The proposed method is able to remove rainy streaks from the input image.
}
\label{fig: deraining}
\end{figure*}

\begin{table*}[!t]
\footnotesize
  \caption{Quantitative comparison using the synthetic rainy dataset~\cite{derain_gan}.}
   \vspace{1mm}
  \label{tab:psnr-derain}
  \centering
  \begin{tabular}{lccccccc}
    \toprule
    Methods &SPM~\cite{derain_spm_tip12}& PRM~\cite{derain_lowrank_iccv13} & CNN~\cite{derain_tip17} & GMM~\cite{Liyu_derain_cvpr16} & ID-CGAN~\cite{derain_gan} & Net-S & Ours\\
    \midrule
    Avg. PSNR &18.88&20.46 & 19.12 & 22.27 & 22.73 & 22.18 &\bf{24.11}\\
    \bottomrule
  \end{tabular}
\end{table*}

\begin{figure*}[!t]\footnotesize
\begin{center}
\begin{tabular}{cccc}
\hspace{-4mm}
\includegraphics[width = 0.24\linewidth]{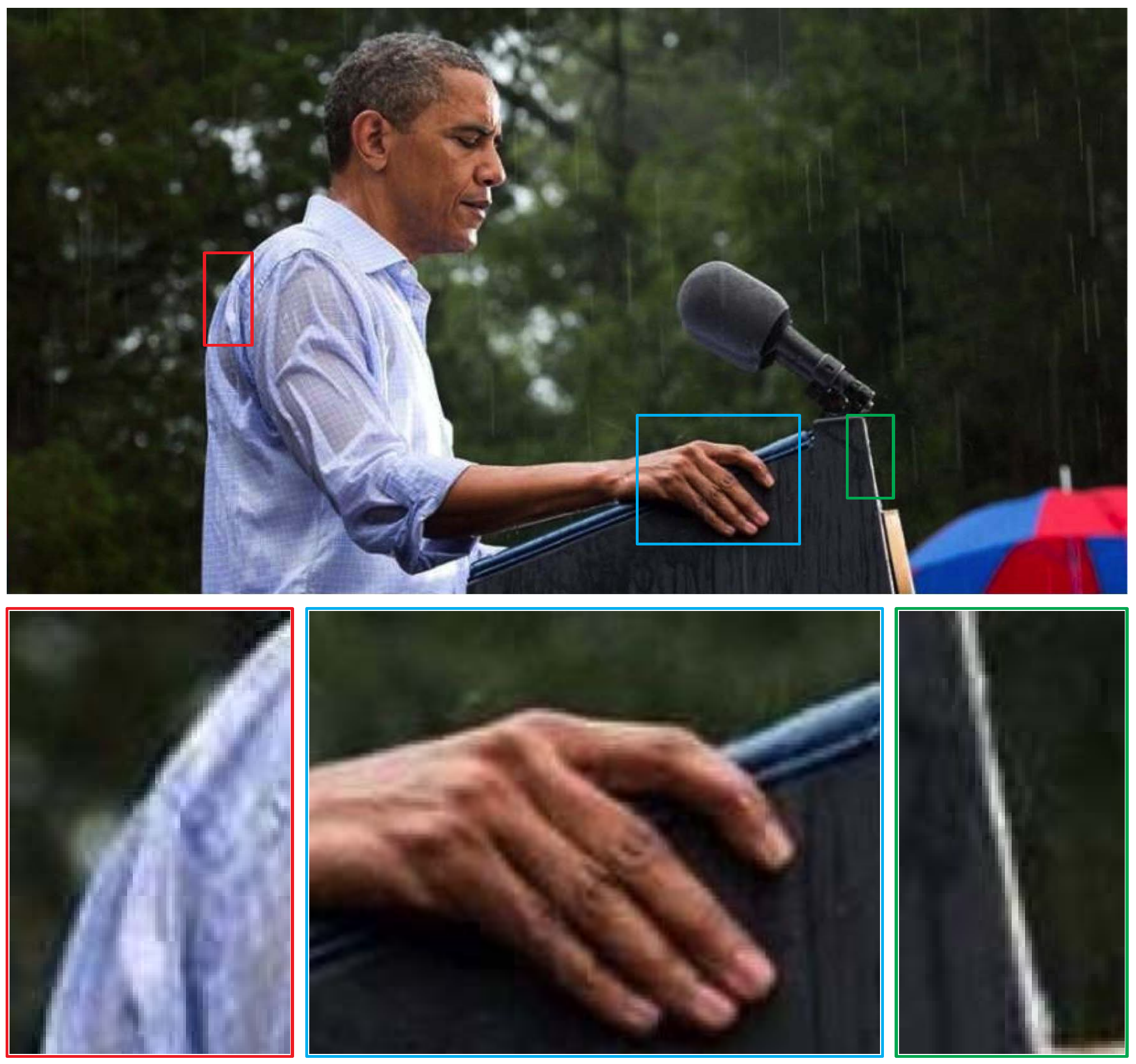}& \hspace{-4mm}
\includegraphics[width = 0.24\linewidth]{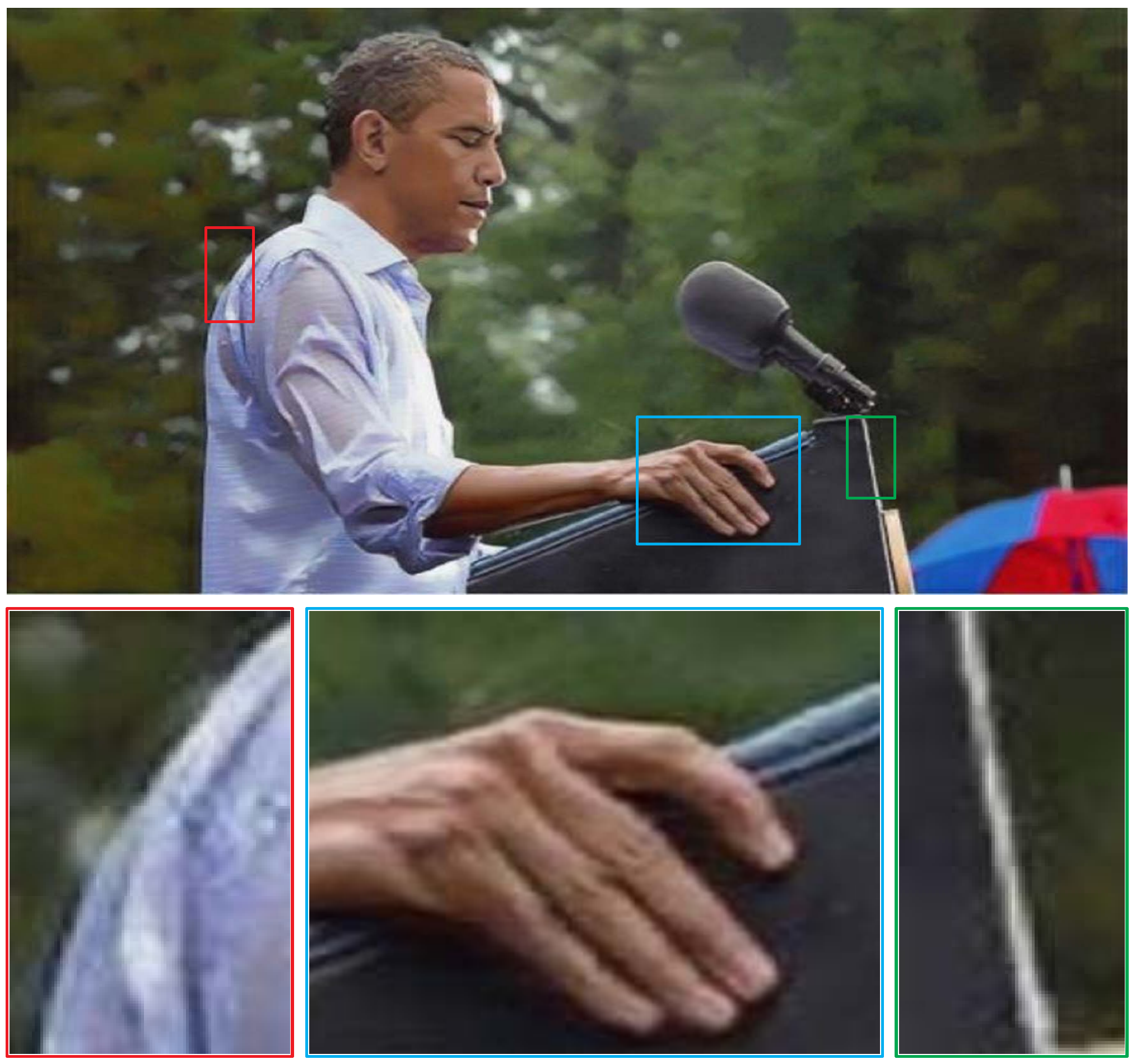}& \hspace{-4mm}
\includegraphics[width = 0.24\linewidth]{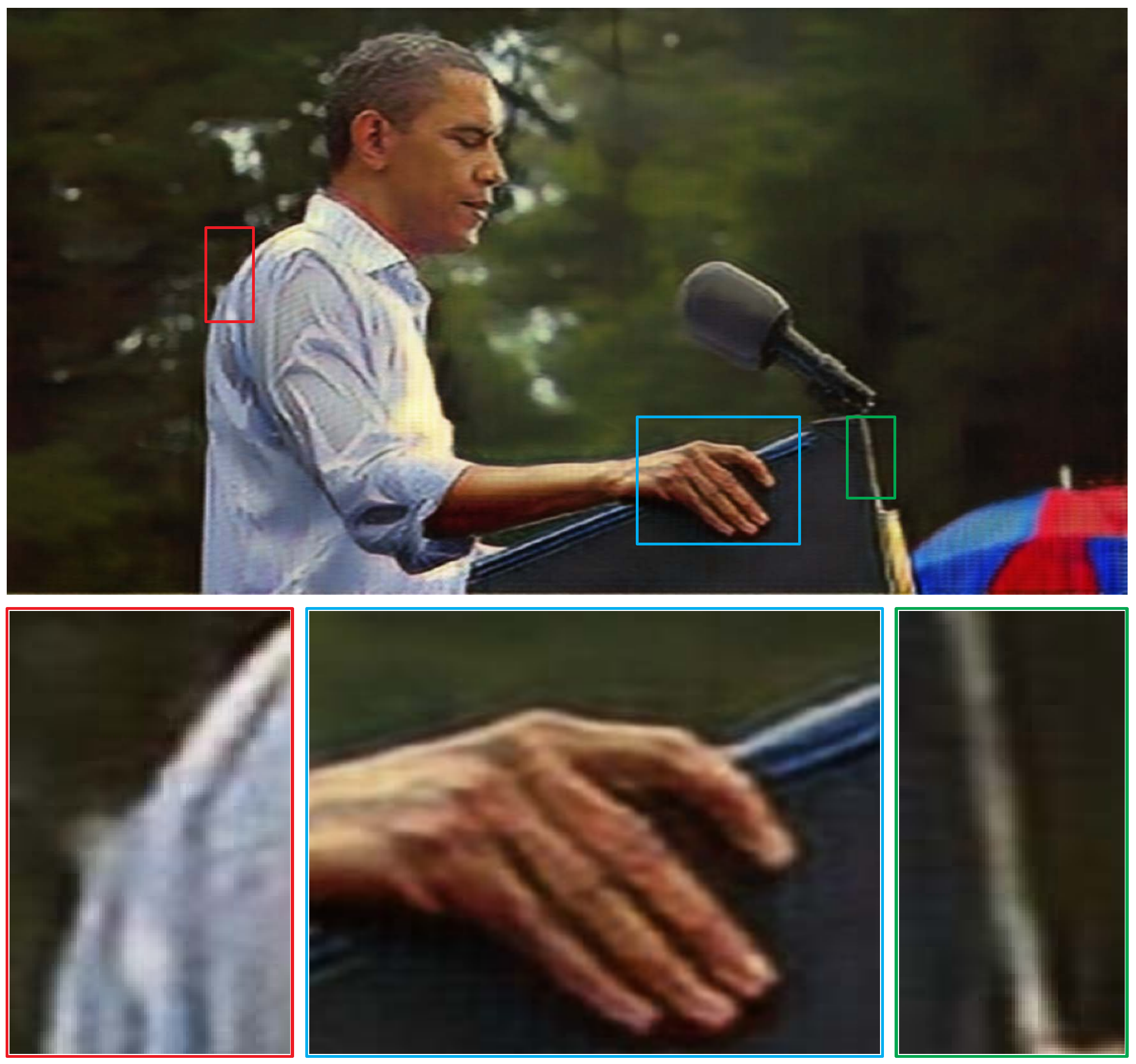}& \hspace{-4mm}
\includegraphics[width = 0.24\linewidth]{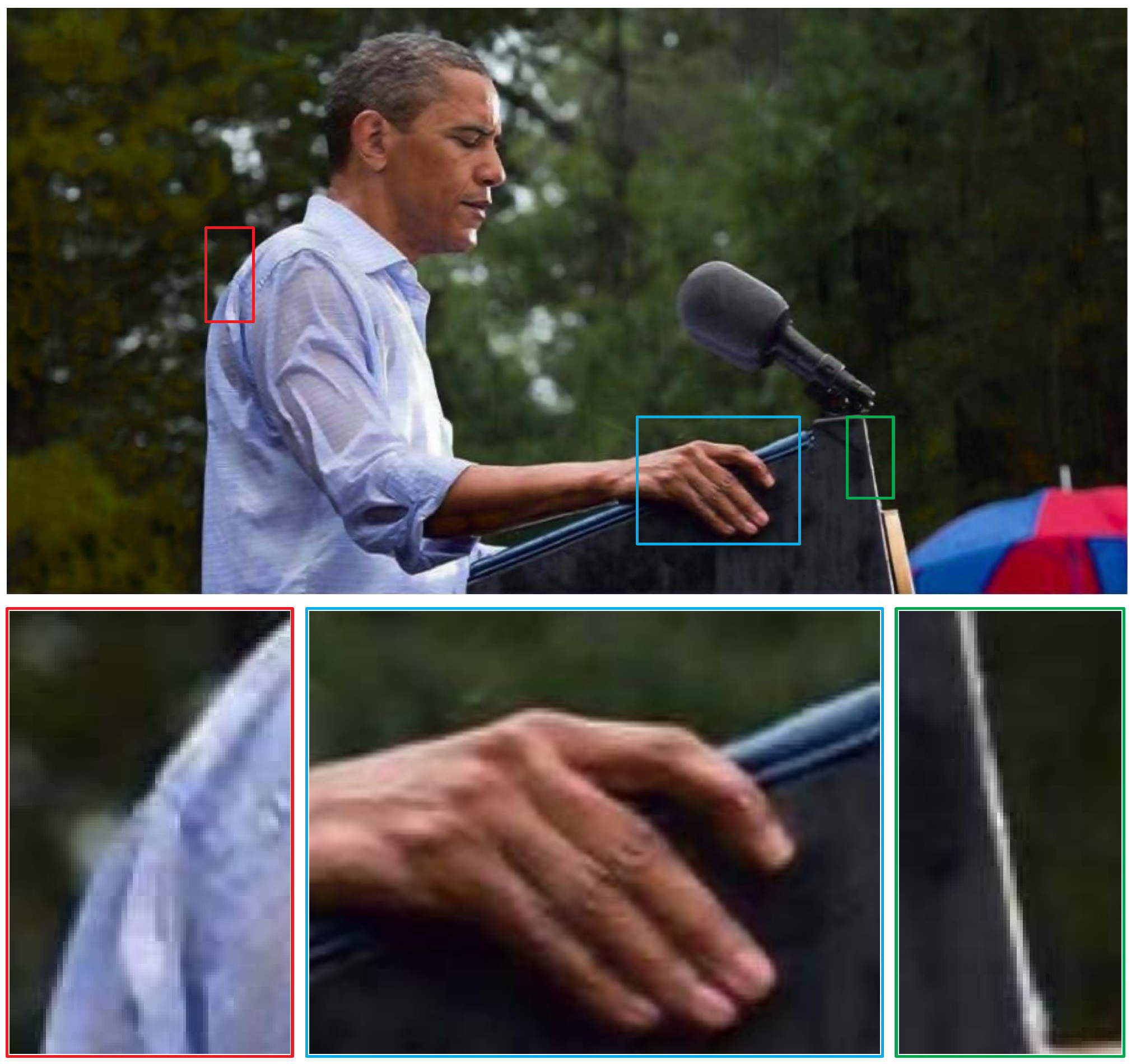}\\
\hspace{-4mm}(a) Input &\hspace{-4mm} (b) ID-CGAN~\cite{derain_gan} &\hspace{-4mm} (c) Fu et al.~\cite{derain_cvpr17_fu} &\hspace{-4mm} (d) Ours \\
\end{tabular}
\end{center}
\vspace{-3mm}
\caption{Visual comparisons of deep learning-based methods for image deraining on real examples. The proposed method is able to remove rainy streaks from the input image and generates much better images with fine details.
}
\label{fig: deraining-real-example}
\end{figure*}

\begin{table*}[!t]
\footnotesize
  \caption{Quantitative comparison using the synthetic hazy image dataset~\cite{dehaze_eccv16}.}
   \vspace{1mm}
  \label{tab:psnr-dehazing}
  \centering
  \begin{tabular}{lcccc}
    \toprule
    Methods &He et al.~\cite{he_dark_channel_dehazing_cvpr09} &Meng et al.~\cite{meng_dehaze_iccv13}&Ren et al.~\cite{dehaze_eccv16} & Ours\\
    \midrule
    Avg. PSNR &15.86&15.06&18.38&\bf{18.85}\\
    \bottomrule
  \end{tabular}
\end{table*}


\begin{figure*}[!t]\footnotesize
\begin{center}
\begin{tabular}{cccc}
\includegraphics[width = 0.22\linewidth, height = 0.26\linewidth]{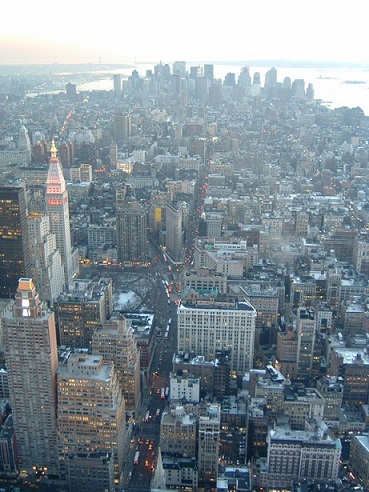}& \hspace{-4mm}
\includegraphics[width = 0.22\linewidth, height = 0.26\linewidth]{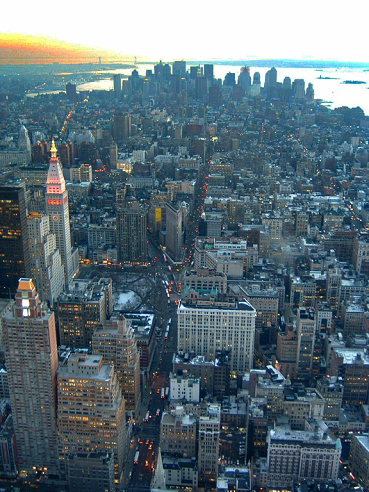}& \hspace{-4mm}
\includegraphics[width = 0.22\linewidth, height = 0.26\linewidth]{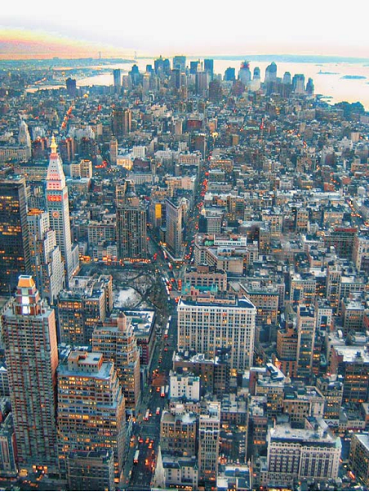}& \hspace{-4mm}
\includegraphics[width = 0.22\linewidth, height = 0.26\linewidth]{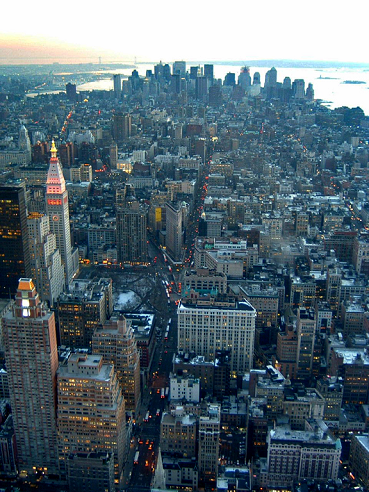}\\
(a) Input & \hspace{-4mm} (b) He et al.~\cite{he_dark_channel_dehazing_cvpr09} & \hspace{-4mm} (c) Tarel et al.~\cite{Tarel_dehazing} & \hspace{-4mm} (d) Cai et al.~\cite{DehazeNet_tip16}\\
\includegraphics[width = 0.22\linewidth, height = 0.26\linewidth]{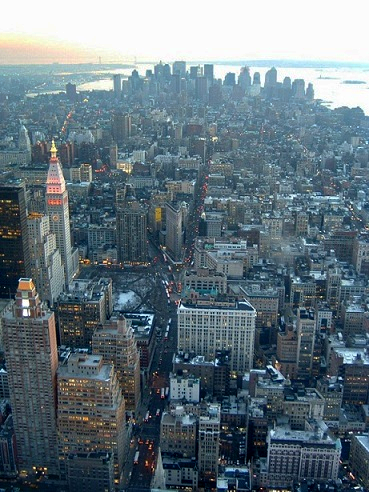} & \hspace{-4mm}
\includegraphics[width = 0.22\linewidth, height = 0.26\linewidth]{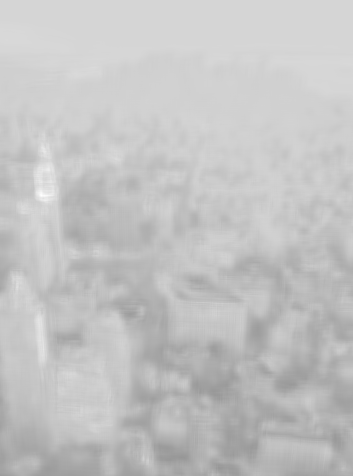}& \hspace{-4mm}
\includegraphics[width = 0.22\linewidth, height = 0.26\linewidth]{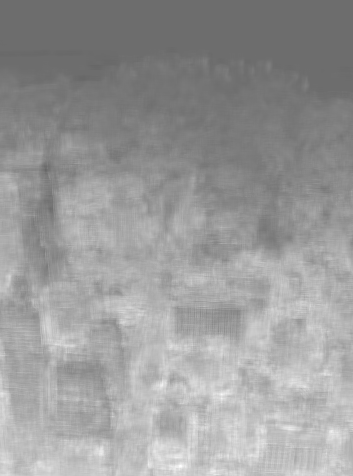}& \hspace{-4mm}
\includegraphics[width = 0.22\linewidth, height = 0.26\linewidth]{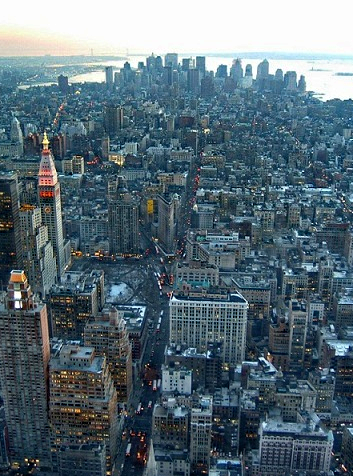}\\
 (d) Ren et al.~\cite{dehaze_eccv16} & \hspace{-4mm} (e) Estimated $S$ & \hspace{-4mm} (f) Estimated $D$ & \hspace{-4mm} (g) Ours\\
\end{tabular}
\end{center}
\vspace{-3mm}
\caption{Visual comparisons for image dehazing. In contrast to the CNN-based methods~\cite{DehazeNet_tip16,dehaze_eccv16} which additionally use conventional methods to estimate atmosphere light, the proposed method directly estimates the atmosphere light in (e) and transmission map in (f) and thus leads to comparable results.
}
\label{fig: dehazing}
\end{figure*}

Figure~\ref{fig: deraining} shows deraining results from the evaluated methods.
The proposed algorithm can accurately estimate both clear details and structures from the input image.
The plain CNN-based methods~\cite{derain_tip17},~\cite{derain_gan} and Net-S all generate results with obvious rainy streaks, demonstrating the advantage of simultaneously recovering structures and details using the DualCNN.

We further evaluate DualCNN using real examples. Figure~\ref{fig: deraining-real-example} shows a real example.
We note that the algorithm in~\cite{derain_cvpr17_fu} develops a deep details network for image deraining. The derained images are obtained by extracting details from input.
However, this method depends on whether the image decomposition method is able to extract details or not.
The results shown in Figure~\ref{fig: deraining-real-example}(c) demonstrate the algorithm in~\cite{derain_cvpr17_fu} fails to generate clearer images.
In contrast, our method generates much clearer results compared to state-of-the-art algorithms.

\vspace{-2mm}
\subsection{Image Dehazing}
\label{ssec: image-dehazing}
\vspace{-2mm}
As discussed in Section~\ref{ssec: generalization}, the proposed method can be applied to the image dehazing.
Similar to the method in~\cite{dehaze_eccv16}, we synthesize the hazy image dataset using the NYU depth dataset~\cite{nyu_depthdata_eccv12} and generate the training data by
randomly sampling 1 million patches including hazy/clear pairs ($I$/$J$),  atmospheric light ($S$), transmission map ($D$).
The size of each image patch used in training stage is $32\times 32$ pixels.
The weights $\alpha$, $\lambda$ and $\gamma$ are set to be 0.1, 0.9, and 0.9, respectively.

We quantitatively evaluate our method on the synthetic hazy images~\cite{dehaze_eccv16}.
As summarized in Table~\ref{tab:psnr-dehazing}, the proposed method performs favorably against the state-of-the-art methods for image dehazing.
The dehazed images in Figure~\ref{fig: dehazing} show that the proposed method can
recover the atmospheric light (Figure~\ref{fig: dehazing}(e)) and transmission map
(Figure~\ref{fig: dehazing}(f)) well, thereby facilitating to recover the clear image (Figure~\ref{fig: dehazing}(g)).


\vspace{-1.5mm}
\section{Analysis and Discussion}
\vspace{-1.5mm}
In this section, we further analyze the proposed method and compare it with the most related methods.

\vspace{-3mm}
{\flushleft \bf{Effect of the architectures of DualCNN.}}
Lin et al.~\cite{bilinear_cnn_iccv15} develop a bilinear model to extract complementary features for fine-grained visual recognition.
%
By contrast, the proposed DualCNN is motivated by the decomposition of a signal into structures and details.
More importantly, the formulation of the proposed model
facilitates incorporating the domain knowledge of each individual application.
Thus, the DualCNN model can be effectively applied to numerous low-level vision problems, e.g., super-resolution, image filtering, deraining, and dehazing.
%
%

\begin{figure*}[!t]\footnotesize
\begin{center}
\begin{tabular}{cccccc}
\includegraphics[width = 0.16\linewidth]{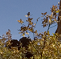}& \hspace{-4mm}
\includegraphics[width = 0.16\linewidth]{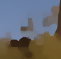}& \hspace{-4mm}
\includegraphics[width = 0.16\linewidth]{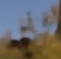}& \hspace{-4mm}
\includegraphics[width = 0.16\linewidth]{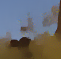}& \hspace{-4mm}
\includegraphics[width = 0.16\linewidth]{figures/9_patch/9_S_patch}& \hspace{-4mm}
\includegraphics[width = 0.16\linewidth]{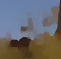}\\
\includegraphics[width = 0.16\linewidth]{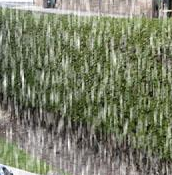}& \hspace{-4mm}
\includegraphics[width = 0.16\linewidth]{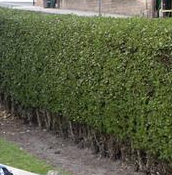}& \hspace{-4mm}
\includegraphics[width = 0.16\linewidth]{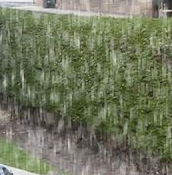}& \hspace{-4mm}
\includegraphics[width = 0.16\linewidth]{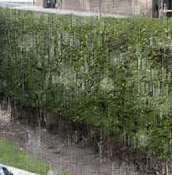}& \hspace{-4mm}
\includegraphics[width = 0.16\linewidth]{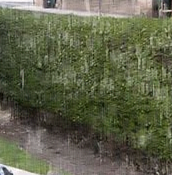}& \hspace{-4mm}
\includegraphics[width = 0.16\linewidth]{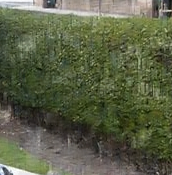}\\
(a) Input &\hspace{-4mm} (b) GT &\hspace{-4mm} (c) SRCNN &\hspace{-4mm} (d) VDSR &\hspace{-4mm} (e) Cascade &\hspace{-4mm} (f) Ours\\
\end{tabular}
\end{center}
\vspace{-3mm}
\caption{Effectiveness of the proposed dual model. (c)-(f) show the comparisons between existing CNNs (including plain net and ResNet) and the proposed net in edge-preserving filtering and image deraining. The plain net (i.e., (c)), ResNet and its deeper version (i.e., (d) and (e)) generate results with significant artifacts. Quantitative evaluations are included in Table~\ref{tab:effect-net}.
}
\label{fig: illustration-filtering}
\end{figure*}
\begin{table}[!t]\footnotesize
  \caption{Quantitative evaluation of different networks on the image filtering~\cite{deepfilter_icml15} and deraining~\cite{derain_gan} datasets in terms of PSNR.}
 \vspace{1mm}
  \label{tab:effect-net}
  \centering
  \begin{tabular}{lccccc}
    \toprule
     Different nets &SRCNN&VDSR&Cascade & Ours\\
    \midrule
    Filtering &32.0&41.6&42.0&\bf{42.1}\\
    Deraining &22.3&23.9&23.5&\bf{24.1}\\
    \bottomrule
  \end{tabular}
\end{table}

\begin{figure*}[!t]\footnotesize
\begin{center}
\begin{tabular}{cc}
\includegraphics[width = 0.98\linewidth]{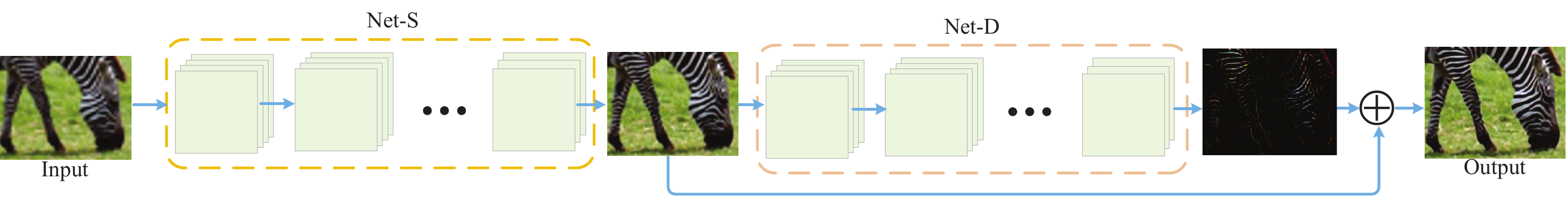}
\end{tabular}
\end{center}
\vspace{-3mm}
\caption{An alternative cascaded architecture that estimates the structure and details sequentially.
}
\label{fig: proposed-model-cascade}
\end{figure*}

Numerous deep learning methods have been developed based on a single branch for low-level vision problems, e.g., SRCNN~\cite{SRCNN_eccv14} and VDSR~\cite{VDSR_cvpr16}.
One natural question is why deeper architectures do not necessarily
lead to better performance.
%
In principle, a sufficiently deep neural network has sufficient capacity to solve any problem
given enough training data.
However, it is non-trivial to learn very deep CNN models for these problems while ensuring high efficiency and simplicity.

For experimental validation, we use the SRCNN and a deeper model, i.e., VDSR, for image filtering and deraining.
The experimental settings are discussed in Section~\ref{sec: Experimental Results}.

Sample results using the VDSR model are shown in Figure~\ref{fig: illustration-filtering}.
While the residual learning (i.e., VDSR) approach performs better than the SRCNN, the generated images with the plain CNN model~\cite{SRCNN_eccv14} contain blurry boundaries or rainy streaks (Figure~\ref{fig: illustration-filtering}(d)).

%

Although the proposed DualCNN consists of two branches, an alternative is to combine the Net-S and Net-D in a cascaded manner as shown in Figure~\ref{fig: proposed-model-cascade}.
In this cascade model, the first stage estimates the main structure while the second stage estimates details.
This network architecture is similar to the ResNet~\cite{ResNet_CVPR16}.
However, this cascaded architecture does not generate high-quality results compared to the proposed DualCNN (Figure~\ref{fig: illustration-filtering}(e) and Table~\ref{tab:effect-net}).

\vspace{-3mm}
{\flushleft \bf{Effect of the loss functions in DualCNN.}}
We evaluate the effects of different loss functions on image dehazing.
Table~\ref{tab:effect-loss} shows that adding
two regularization losses $\mathcal{L}_s$ in~\eqref{eq: loss-function-s} and
$\mathcal{L}_d$ in \eqref{eq: loss-function-t}
significantly improves the performance.

\begin{table}[!t]
\footnotesize
  \caption{Quantitative evaluation of the proposed dual composition loss function on the validation data of image dehazing in terms of PSNR and SSIM.
  }
  \vspace{1mm}
%
  \label{tab:effect-loss}
  \centering
  \begin{tabular}{lccccc}
    \toprule
     ($\lambda/\alpha$, $\gamma/\alpha$) &(0, 0)& (1, 0) & (0, 1) & (9, 9)\\
    \midrule
    Avg. PSNR &21.13& 26.27 & 26.45 & 26.43 \\
    Avg. SSIM &0.7449 & 0.8987  & 0.9139  & 0.9108\\
    \bottomrule
  \end{tabular}
\end{table}

\vspace{-3mm}
{\flushleft \bf{Different architectures in DualCNN.}}
We have used different network structures for the two branches of DualCNNs in the experiments in Section~\ref{sec: Experimental Results}.
It is interesting to test using the same structures for the two branches of a DualCNN.
To this end, we set the two branches in a DualCNN using the network structures of SRCNN~\cite{SRCNN/eccv14} and train the DualCNN according to the
same settings used in the image super-resolution experiment.
%
The trained DualCNN generates the results with higher average PSNR/SSIM values (30.3690/0.8603) than those of SRCNN (30.1496/0.8551) for $\times 4$ upsampling on the ``Set5" dataset.

We further quantitatively evaluate the DualCNN when the two branches are the same on image deraining using synthetic rainy dataset~\cite{derain_gan}. Similar to the image super-resolution experimental settings,
the two branches in the DualCNN are set to be the network structures of SRCNN~\cite{SRCNN_eccv14} (SDCNN-S) and the network structures of VDSR~\cite{VDSR_cvpr16} (SDCNN-D), respectively.
Table~\ref{tab:psnr-derain-structures} shows that DualCNN with deeper model generates better results when the architectures of two branches are the same.
However, the DualCNN where one branch is SRCNN and the other one is VDSR performs better than SDCNN-D.
This is mainly because the main structures of the input images are similar to those of output images. Deeper model used in ``net-S" will introduce errors in the learning stage.

\vspace{-1.5mm}
\begin{table}[!t]
\footnotesize
  \caption{Quantitative evaluation of two branches in DualCNN using the synthetic rainy dataset~\cite{derain_gan}.}
  \vspace{1mm}
  \label{tab:psnr-derain-structures}
  \centering
  \begin{tabular}{lccc}
    \toprule
    Two branches &SDCNN-S& SDCNN-D & Ours\\
    \midrule
    Avg. PSNR &22.42&23.58 &\bf{24.11}\\
    \bottomrule
  \end{tabular}
  \vspace{-3.6mm}
\end{table}

\begin{figure}[!t]\footnotesize
\begin{center}
\begin{tabular}{cc}
\includegraphics[width = 0.8\linewidth]{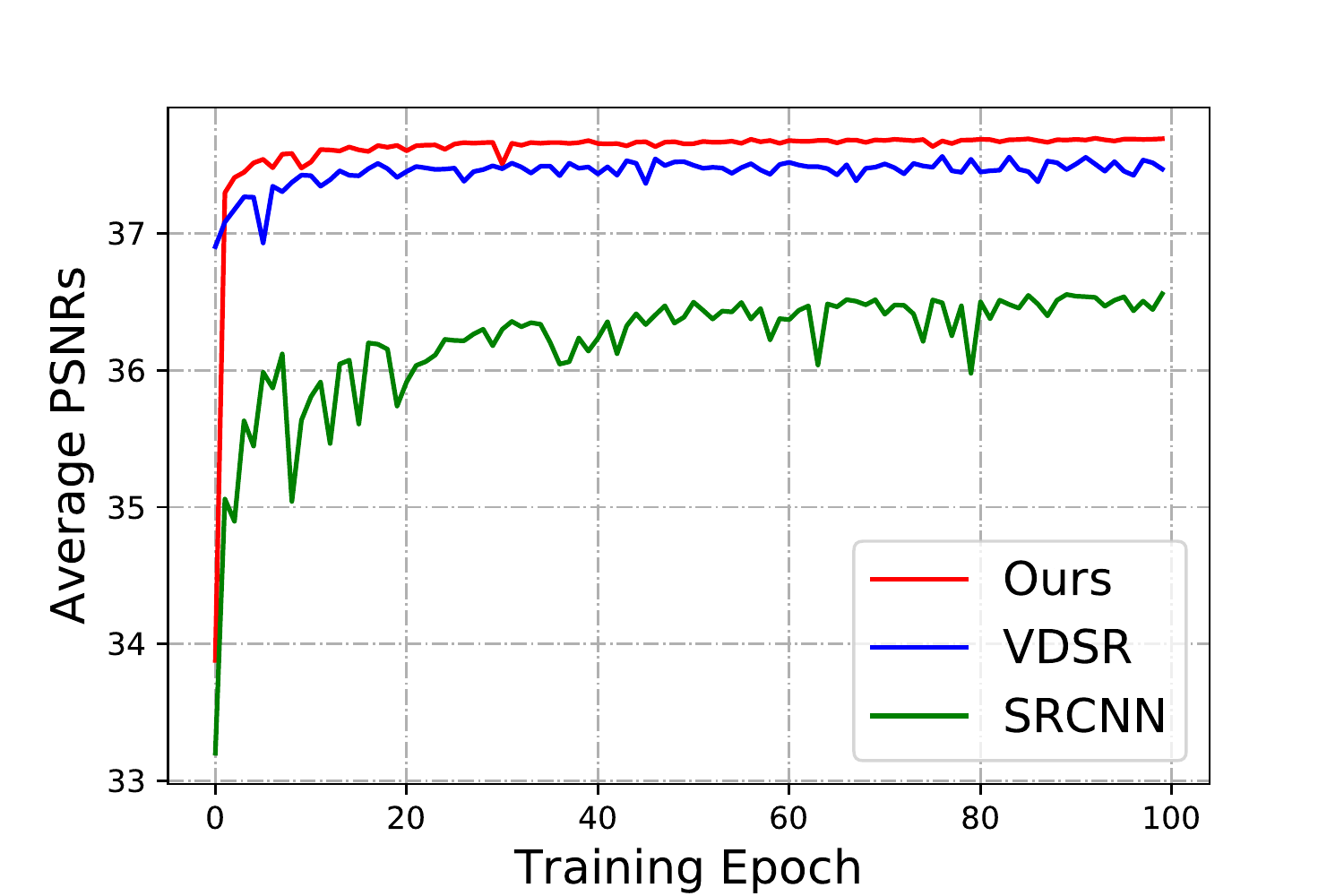}\\
\end{tabular}
\end{center}
\vspace{-0.3cm}
\caption{
Quantitative evaluation of the convergence property on the super-resolution dataset (Set5, $\times 2$).
}
\label{fig: convergence-property}
\end{figure}

\vspace{-2mm}
{\flushleft {\bf Convergence property.}}  We quantitatively evaluate convergence properties of our method on the super-resolution dataset, i.e., Set5.
Although the proposed network contains two branches compared to other methods~\cite{SRCNN_eccv14,VDSR_cvpr16},
it has the similar convergency property to the SRCNN and VDSR as shown in Figure~\ref{fig: convergence-property}.
%

%

\vspace{-3mm}
\section{Conclusion}
\vspace{-2mm}
In this paper, we propose a novel dual convolutional neural network
for low-level vision tasks, called DualCNN.
From an input signal, the DualCNN recovers both the structure and detail components,
which can generate the target signal according to the problem formulation
for a specific task.
We analyze the effect of the DualCNN and show
that it is a generic framework and can be effectively and efficiently applied to
numerous low-level vision tasks,
including image super-resolution, filtering, image deraining, and image dehazing.
Experimental results show that the DualCNN performs favorably against state-of-the-art methods that have been specially designed
for each task.

{\small
	\bibliographystyle{ieee}
	\bibliography{filter}

\begin{thebibliography}{10}\itemsep=-1pt

\bibitem{Burger-CVPR12}
H.~Burger, C.~Schuler, and S.~Harmeling.
\newblock Image denosing: {Can} plain neural networks compete with {BM3D}.
\newblock In {\em CVPR}, 2012.

\bibitem{DehazeNet_tip16}
B.~Cai, X.~Xu, K.~Jia, C.~Qing, and D.~Tao.
\newblock Dehazenet: An end-to-end system for single image haze removal.
\newblock {\em {IEEE} TIP}, 25(11):5187--5198, 2016.

\bibitem{derain_lowrank_iccv13}
Y.-L. Chen and C.-T. Hsu.
\newblock A generalized low-rank appearance model for spatio-temporally
  correlated rain streaks.
\newblock In {\em ICCV}, pages 1968--1975, 2013.

\bibitem{dong_artifacts_iccv15}
C.~Dong, Y.~Deng, C.~C. Loy, and X.~Tang.
\newblock Compression artifacts reduction by a deep convolutional network.
\newblock In {\em ICCV}, pages 576--584, 2015.

\bibitem{SRCNN_eccv14}
C.~Dong, C.~C. Loy, K.~He, and X.~Tang.
\newblock Learning a deep convolutional network for image super-resolution.
\newblock In {\em ECCV}, pages 184--199, 2014.

\bibitem{SRCNN_pami16}
C.~Dong, C.~C. Loy, K.~He, and X.~Tang.
\newblock Image super-resolution using deep convolutional networks.
\newblock {\em {IEEE} TPAMI}, 38(2):295--307, 2016.

\bibitem{fast_SRCNN_eccv16}
C.~Dong, C.~C. Loy, and X.~Tang.
\newblock Accelerating the super-resolution convolutional neural network.
\newblock In {\em ECCV}, pages 391--407, 2016.

\bibitem{Eigen_derain}
D.~Eigen, D.~Krishnan, and R.~Fergus.
\newblock Restoring an image taken through a window covered with dirt or rain.
\newblock In {\em ICCV}, pages 633--640, 2013.

\bibitem{derain_tip17}
X.~Fu, J.~Huang, X.~Ding, Y.~Liao, and J.~Paisley.
\newblock Clearing the skies: A deep network architecture for single-image rain
  removal.
\newblock {\em {IEEE} Trans. Image Processing}, 26(6):2944--2956, 2017.

\bibitem{derain_cvpr17_fu}
X.~Fu, J.~Huang, D.~Zeng, Y.~Huang, X.~Ding, and J.~Paisley.
\newblock Removing rain from single images via a deep detail network.
\newblock In {\em CVPR}, pages 3855--3863, 2017.

\bibitem{RCNN_fast_iccv15}
R.~B. Girshick.
\newblock Fast {R-CNN}.
\newblock In {\em ICCV}, pages 1440--1448, 2015.

\bibitem{he_dark_channel_dehazing_cvpr09}
K.~He, J.~Sun, and X.~Tang.
\newblock Single image haze removal using dark channel prior.
\newblock In {\em CVPR}, pages 1956--1963, 2009.

\bibitem{ResNet_CVPR16}
K.~He, X.~Zhang, S.~Ren, and J.~Sun.
\newblock Deep residual learning for image recognition.
\newblock In {\em CVPR}, pages 770--778, 2016.

\bibitem{SelfEx_sr}
J.-B. Huang, A.~Singh, and N.~Ahuja.
\newblock Single image super-resolution from transformed self-exemplars.
\newblock In {\em CVPR}, pages 5197--5206, 2015.

\bibitem{denoising_deep_learning_nips08}
V.~Jain and H.~S. Seung.
\newblock Natural image denoising with convolutional networks.
\newblock In {\em NIPS}, pages 769--776, 2008.

\bibitem{derain_spm_tip12}
L.-W. Kang, C.-W. Lin, and Y.-H. Fu.
\newblock Automatic single-image-based rain streaks removal via image
  decomposition.
\newblock {\em {IEEE} TIP}, 21(4):1742--1755, 2012.

\bibitem{VDSR_cvpr16}
J.~Kim, J.~K. Lee, and K.~M. Lee.
\newblock Accurate image super-resolution using very deep convolutional
  networks.
\newblock In {\em CVPR}, pages 1646--1654, 2016.

\bibitem{Recurrent_SR_cvpr16}
J.~Kim, J.~K. Lee, and K.~M. Lee.
\newblock Deeply-recursive convolutional network for image super-resolution.
\newblock In {\em CVPR}, pages 1637--1645, 2016.

\bibitem{Alexnet_nips12}
A.~Krizhevsky, I.~Sutskever, and G.~E. Hinton.
\newblock Imagenet classification with deep convolutional neural networks.
\newblock In {\em NIPS}, pages 1106--1114, 2012.

\bibitem{SRGAN}
C.~Ledig, L.~Theis, F.~Huszar, J.~Caballero, A.~Cunningham, A.~Acosta,
  A.~Aitken, A.~Tejani, J.~Totz, Z.~Wang, and W.~Shi.
\newblock Photo-realistic single image super-resolution using a generative
  adversarial network.
\newblock In {\em CVPR}, pages 4681--4690, 2017.

\bibitem{Liyu_derain_cvpr16}
Y.~Li, R.~T. Tan, X.~Guo, J.~Lu, and M.~S. Brown.
\newblock Rain streak removal using layer priors.
\newblock In {\em CVPR}, pages 2736--2744, 2016.

\bibitem{videoSR_iccv15}
R.~Liao, X.~Tao, R.~Li, Z.~Ma, and J.~Jia.
\newblock Video super-resolution via deep draft-ensemble learning.
\newblock In {\em ICCV}, pages 531--539, 2015.

\bibitem{bilinear_cnn_iccv15}
T.-Y. Lin, A.~{Roy Chowdhury}, and S.~Maji.
\newblock Bilinear {CNN} models for fine-grained visual recognition.
\newblock In {\em ICCV}, pages 1449--1457, 2015.

\bibitem{rnnfilter_eccv16}
S.~Liu, J.~Pan, and M.-H. Yang.
\newblock Learning recursive filters for low-level vision via a hybrid neural
  network.
\newblock In {\em ECCV}, pages 560--576, 2016.

\bibitem{BSDS}
D.~R. Martin, C.~C. Fowlkes, D.~Tal, and J.~Malik.
\newblock A database of human segmented natural images and its application to
  evaluating segmentation algorithms and measuring ecological statistics.
\newblock In {\em ICCV}, pages 416--425, 2001.

\bibitem{meng_dehaze_iccv13}
G.~Meng, Y.~Wang, J.~Duan, S.~Xiang, and C.~Pan.
\newblock Efficient image dehazing with boundary constraint and contextual
  regularization.
\newblock In {\em ICCV}, pages 617--624, 2013.

\bibitem{sijieren_sr_nips15}
J.~S.~J. Ren, L.~Xu, Q.~Yan, and W.~Sun.
\newblock Shepard convolutional neural networks.
\newblock In {\em NIPS}, pages 901--909, 2015.

\bibitem{dehaze_eccv16}
W.~Ren, S.~Liu, H.~Zhang, J.~Pan, X.~Cao, and M.-H. Yang.
\newblock Single image dehazing via multi-scale convolutional neural networks.
\newblock In {\em ECCV}, pages 154--169, 2016.

\bibitem{ESPCN}
W.~Shi, J.~Caballero, F.~Huszar, J.~Totz, A.~P. Aitken, R.~Bishop, D.~Rueckert,
  and Z.~Wang.
\newblock Real-time single image and video super-resolution using an efficient
  sub-pixel convolutional neural network.
\newblock In {\em CVPR}, pages 1874--1883, 2016.

\bibitem{nyu_depthdata_eccv12}
N.~Silberman, D.~Hoiem, P.~Kohli, and R.~Fergus.
\newblock Indoor segmentation and support inference from {RGBD} images.
\newblock In {\em ECCV}, pages 746--760, 2012.

\bibitem{deep_face_application1}
Y.~Sun, Y.~Chen, X.~Wang, and X.~Tang.
\newblock Deep learning face representation by joint
  identification-verification.
\newblock In {\em NIPS}, pages 1988--1996, 2014.

\bibitem{Tarel_dehazing}
J.~Tarel, N.~Hauti{\`{e}}re, L.~Caraffa, A.~Cord, H.~Halmaoui, and D.~Gruyer.
\newblock Vision enhancement in homogeneous and heterogeneous fog.
\newblock {\em {IEEE} Intell. Transport. Syst. Mag.}, 4(2):6--20, 2012.

\bibitem{A+_accv15}
R.~Timofte, V.~D. Smet, and L.~J.~V. Gool.
\newblock {A+:} adjusted anchored neighborhood regression for fast
  super-resolution.
\newblock In {\em ACCV}, pages 111--126, 2014.

\bibitem{denoising_deep_learning_nips12}
J.~Xie, L.~Xu, and E.~Chen.
\newblock Image denoising and inpainting with deep neural networks.
\newblock In {\em NIPS}, pages 350--358, 2012.

\bibitem{xul0_smooth_tog11}
L.~Xu, C.~Lu, Y.~Xu, and J.~Jia.
\newblock Image smoothing via \emph{L}\({}_{\mbox{0}}\) gradient minimization.
\newblock {\em {ACM} TOG}, 30(6):174:1--174:12, 2011.

\bibitem{deepfilter_icml15}
L.~Xu, J.~S.~J. Ren, Q.~Yan, R.~Liao, and J.~Jia.
\newblock Deep edge-aware filters.
\newblock In {\em ICML}, pages 1669--1678, 2015.

\bibitem{rtv_sa12}
L.~Xu, Q.~Yan, Y.~Xia, and J.~Jia.
\newblock Structure extraction from texture via relative total variation.
\newblock {\em {ACM} TOG}, 31(6):139:1--139:10, 2012.

\bibitem{derain_gan}
H.~Zhang, V.~Sindagi, and V.~M. Patel.
\newblock Image de-raining using a conditional generative adversarial network.
\newblock {\em CoRR}, abs/1701.05957, 2017.

\bibitem{wmf}
Q.~Zhang, L.~Xu, and J.~Jia.
\newblock 100+ times faster weighted median filter {(WMF)}.
\newblock In {\em CVPR}, pages 2830--2837, 2014.

\end{thebibliography}
}

\end{document}